%% file: arxiv_deepthink.tex
\documentclass[11pt,letterpaper]{article}

\usepackage{amsmath,amssymb,amsthm}
\usepackage{deepthink}

\usepackage[nameinlink]{cleveref}
\input{macro_deepthink}

\usepackage{color}
\usepackage{dsfont}
\usepackage[overload]{empheq}

\usepackage{hyperref}       
\usepackage{algorithm}
\usepackage{algpseudocode}
\usepackage{colortbl}
\usepackage{bm}
\definecolor{Gray}{gray}{0.85}
\usepackage[nameinlink]{cleveref}
\usepackage[
  backend=biber,
  style=alphabetic,
  maxbibnames=100,
  minbibnames=100,
  maxcitenames=2,
  mincitenames=2
]{biblatex}
\usepackage{url}            
\usepackage{booktabs}       

\usepackage{graphicx}
\usepackage{makecell}

\newcommand*{\Scale}[2][4]{\scalebox{#1}{$#2$}}%
\usepackage{tcolorbox}

\title{Improving Efficiency of Diffusion Models via Multi-Stage Framework and Tailored Multi-Decoder Architectures}

\newcommand{\jointfirst}{\textsuperscript{\dag}}
\newcommand{\corrauth}{\textsuperscript{\ddag}}

\authorblock{
  \href{https://www.huijiezh.com/}{Huijie Zhang}\jointfirst \textsuperscript{1},
  \href{https://scholar.google.com/citations?user=ybsmKpsAAAAJ&hl=en}{Yifu Lu}\jointfirst \textsuperscript{1},
  \href{https://sites.google.com/view/ismailalkhouri/about}{Ismail Alkhouri}\textsuperscript{1, 2},
  \href{https://sites.google.com/site/sairavishankar3/}{Saiprasad Ravishankar} \textsuperscript{2},
  \href{https://statistics.ucdavis.edu/people/dogyoon-song}{Dogyoon Song} \textsuperscript{1},
  \href{https://qingqu.engin.umich.edu/}{Qing Qu}\corrauth \textsuperscript{1}
}

\affiliation{
  \textsuperscript{1} University of Michigan 
  \textsuperscript{2} Michigan State University 
}

\authornote{
  \jointfirst\ Joint first author \quad
  \corrauth\ Corresponding author
}

\abstracttext{
\noindent Diffusion models are powerful generative frameworks but suffer from slow training and sampling due to long diffusion trajectories and large time-conditioned networks. We propose a multi-stage diffusion framework motivated by empirical evidence that combining timestep-specific parameters with universally shared representations improves efficiency. Our method partitions the diffusion timeline into stages and employs a multi-decoder U-Net with a shared encoder and stage-specific decoders, enabling better allocation of computational resources and reducing inter-stage interference. Extensive experiments on three state-of-the-art diffusion models, including large-scale latent diffusion models, demonstrate substantial improvements in training and sampling efficiency.
}


\date{\today}
\correspondence{\href{mailto:huijiezh@umich.edu}{huijiezh@umich.edu}}
\resources{\quad \href{https://www.huijiezh.com/multistage-diffusion-model/}{Project page} \quad $\cdot$ \quad \href{https://github.com/LucasYFL/Multistage_Diffusion}{Code}}

\headerlogo{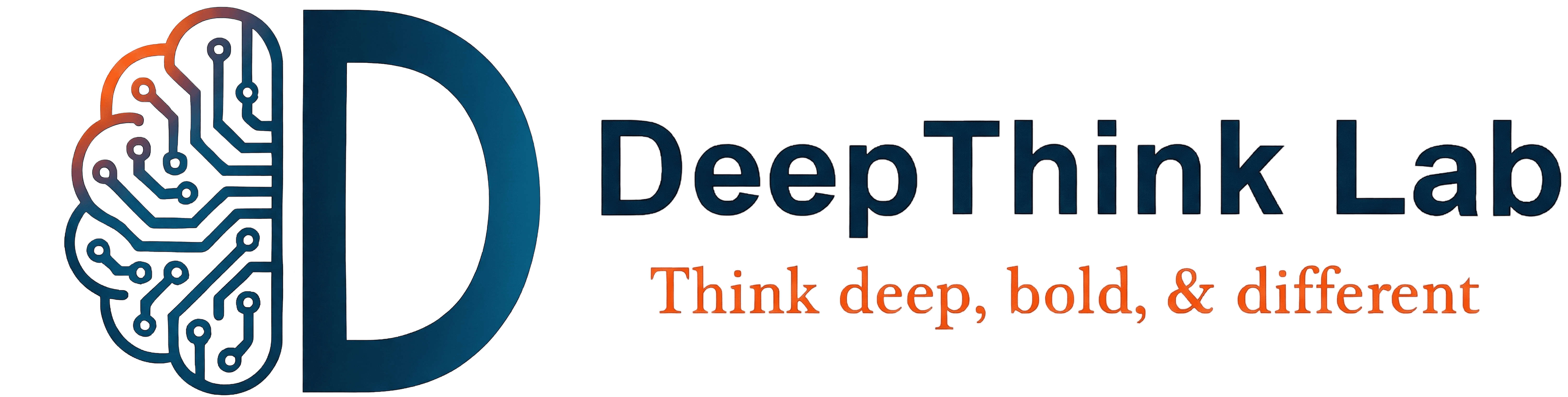}{https://deepthink-umich.github.io}

\begin{document}

\makeDeepthinkHeader


\begin{figure}[h]
\centering
\captionsetup{type=figure}
\includegraphics[width=\linewidth]{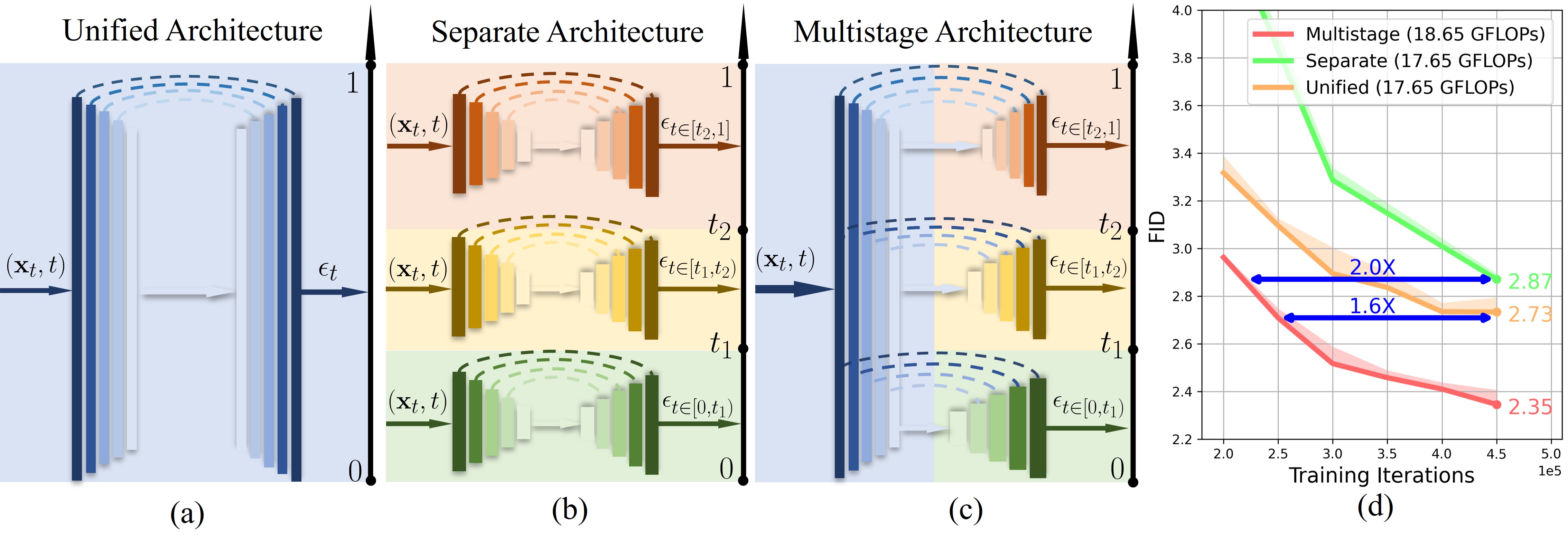}
\caption{\textbf{Overview of three diffusion model architectures:} (a) unified, (b) separate, and (c) our proposed multistage architectures. Compared with (a) and (b), our approach improves sampling quality, and significantly enhances training efficiency, as indicated by the FID scores and their corresponding training iterations (d).}
\label{fig:arch}
\end{figure}

\newpage
\tableofcontents

\newpage

\input{section/intro_arxiv}

\input{section/relatedwork}
\input{section/method}

\input{section/analysis}
\input{section/experiments_arxiv}

\input{section/conclusion}

\section*{Acknowledgment} 
HJZ, YFL, and QQ acknowledge support from NSF CAREER CCF-2143904, NSF CCF-2212066, NSF CCF-2212326, NSF IIS 2312842, ONR N00014-22-1-2529, an AWS AI Award, and a gift grant from KLA. QQ also acknowledges the support from MICDE Catalyst Grant. IA and SR acknowledge the support from the NSF grants CCF-2212065 and CCF-2212066. Results presented in this paper were obtained using CloudBank, which is supported by the NSF under Award \#1925001, and the authors acknowledge efficient cloud management framework SkyPilot \cite{yang2023skypilot} for computing. The authors acknowledge valuable discussions with Prof. Jeffrey Fessler (U. Michigan), Prof. Liyue Shen (U. Michigan), and Prof. Rongrong Wang (MSU).

\newpage 

\printbibliography
\newpage
\appendix

\input{section/appendix_theorem_arxiv}
\input{section/appendix_threestage_split_arxiv}
\input{section/appendix_more_exp_arxiv}

\end{document}

%% file: macro_deepthink.tex
\usepackage{tgpagella}
\usepackage[utf8]{inputenc} 
\usepackage[T1]{fontenc}    
\usepackage{url}
\usepackage{amsmath,amsthm,amssymb,amsbsy,amsfonts,amscd,bm}
\usepackage{xcolor}
\usepackage{color}
\usepackage{graphicx}
\graphicspath{{./figs/}}
\usepackage{algorithm}
\usepackage{comment}
\usepackage{multirow}
\usepackage{fancyhdr}
\usepackage{authblk}
\usepackage{cleveref}
\usepackage{subcaption}
\usepackage{wrapfig}
\usepackage{bbding}
\usepackage{algpseudocode}
\usepackage{graphicx}
\usepackage{mathtools}
\usepackage{multirow}
\usepackage{cleveref}
\usepackage{tcolorbox}
\usepackage{bbding}
\usepackage{subcaption}
\usepackage{wrapfig}
\usepackage{titletoc}
\usepackage{titlesec}
\usepackage{amsthm}

\usepackage[toc, page]{appendix}

\newtheorem{prop}{Proposition}

\theoremstyle{remark}

\newcommand{\e}{\begin{equation}}
\newcommand{\ee}{\end{equation}}
\newcommand{\en}{\begin{equation*}}
\newcommand{\een}{\end{equation*}}
\newcommand{\eqn}{\begin{eqnarray}}
\newcommand{\eeqn}{\end{eqnarray}}
\newcommand{\bmat}{\begin{bmatrix}}
\newcommand{\emat}{\end{bmatrix}}

\DeclareMathAlphabet\mathbfcal{OMS}{cmsy}{b}{n}

\DeclareMathOperator*{\argmin}{\text{arg~min}}
\DeclareMathOperator*{\argmax}{\text{arg~max}}

\newcommand{\eps}{\epsilon}

 \newcommand{ \Brac }[1]{\left\lbrace #1 \right\rbrace}
 \newcommand{ \brac }[1]{\left[ #1 \right]}

\graphicspath{{./figs/}}

\newlength{\imgwidth}
\newboolean{twoColVersion}
\setboolean{twoColVersion}{false}
\newcommand{\twoCol}[2]{\ifthenelse{\boolean{twoColVersion}} {#1} {#2} }

\long\def\comment#1{}

\long\def\red#1{\bgroup\color{red}#1\egroup}

\definecolor{mich-blue}{HTML}{0027CC}
\definecolor{mich-blue-high}{HTML}{0027CC}
\definecolor{red-high}{HTML}{CA2020}
\definecolor{green-high}{HTML}{20A520}
\definecolor{mich-maize}{HTML}{FFCB05}
\definecolor{law-stone}{HTML}{655A52}
\definecolor{burton-beige}{HTML}{9B9A9D}
\definecolor{arch-ivy}{HTML}{7E732F}

 \colorlet{color1}{gray!15}

%% file: section/intro_arxiv.tex
\section{Introduction}



Recently, diffusion models have emerged as powerful deep generative modeling tools, showcasing remarkable performance in various applications, ranging from unconditional image generation \cite{ho2020denoising, song2020score}, conditional image generation \cite{dhariwal2021diffusion, ho2022classifier}, image-to-image translation \cite{su2022dual, saharia2022palette, zhao2022egsde}, text-to-image generation \cite{rombach2022high, ramesh2021zero, nichol2021glide}, inverse problem solving \cite{song2023solving, chung2022diffusion, song2021solving, alkhouri2023diffusion}, video generation \cite{ho2022imagen, harvey2022flexible}, and so on.
These models employ a training process involving continuous injection of noise into training samples (``diffusion''), which are then utilized to generate new samples, such as images, by transforming random noise instances through a reverse diffusion process guided by the ``score function\footnote{The (Stein) score function of distribution $p_t$ at $\bm{x}_t$ is $\nabla_{\bm{x}_t} \log p_t(\bm{x}_t)$. 
}'' of the data distribution learned by the model. Moreover, recent work demonstrates that those diffusion models enjoy optimization stability and model reproducibility compared with other types of generative models \cite{zhang2023emergence}.
However, diffusion models suffer from slow training and sampling despite their remarkable generative capabilities, which hinders their use in applications where real-time generation is desired \cite{ho2020denoising, song2020score}. These drawbacks primarily arise from the necessity of tracking extensive forward and reverse diffusion trajectories, as well as managing a large model with numerous parameters across multiple timesteps (i.e., diffusion noise levels).

In this paper, we address these challenges based on two key observations: (i) there exists substantial parameter redundancy in current diffusion models, and (ii) they are trained inefficiently due to dissimilar gradients across different noise levels. Specifically, we find that 
training diffusion models require fewer parameters to accurately learn the score function at high noise levels, while larger parameters are needed at low noise levels. 
Furthermore, we also observe that when learning the score function, distinct shapes of distributions at different noise levels result in dissimilar gradients, which appear to slow down the training process driven by gradient descent.



Building on these insights, we propose a multi-stage framework with two key components: (1) a multi-decoder U-net architecture, and (2) a new partitioning algorithm to cluster timesteps (noise levels) into distinct stages. 
In terms of our new architecture, we design a multi-decoder U-Net that incorporates one universal encoder shared across all intervals and individual decoders tailored to each time stage; see \Cref{fig:arch} (c) for an illustration. This approach combines the advantages of both universal and stage-specific architectures, which is much more efficient that using a single architecture for the entire training process \cite{song2020score, ho2020denoising, karras2022elucidating} (\Cref{fig:arch} (a)). Moreover, compared to previous approaches that completely separate architectures for each sub-interval \cite{choi2022perception, go2023addressing, lee2023multi, go2023towards} (\Cref{fig:arch} (b)), our approach can effectively mitigate interference between stages arising from disparate gradient effects, leading to improved efficiency. On the other hand, when it comes to partitioning the training stages of our network, we designed an algorithm aimed at grouping the timesteps. This is achieved by minimizing the functional distance within each cluster in the training objective and making use of the optimal denoiser formulation \cite{karras2022elucidating}. By integrating these two key components, our framework enables efficient allocation of computational resources (e.g., U-net parameters) and stage-tailored parameterization. In \Cref{sec:experiments}, we showcase via extensive numerical experiments that our framework effectively improves both training and sampling efficiency. These experiments are performed on diverse benchmark datasets, demonstrating significant acceleration by using our framework when compared to three state-of-the-art (SOTA) diffusion model architectures.


\paragraph{Contributions.} As a summary, the major contributions of this work can be highlighted as follows: 
\begin{itemize}[leftmargin=*]
    \item \textbf{Identifying two key sources of inefficiency.}
    We identified two key sources that cause inefficiencies in training diffusion models across various time step: (1) a significant variation in the requirement of model capacity, and (2) the gradient dissimilarity. As such, using a unified network cannot meet with the changing requirement at different time steps.
   \item \textbf{A new multi-stage framework.}
    We introduced a new multi-stage architecture, illustrated in \Cref{fig:arch} (c). We tackle the challenges by segmenting the time interval into multiple stages, where we employ customized multi-decoder U-net architectures that blends time-dependent models with a universally shared encoder.
    \item \textbf{Improved training and sampling efficiency.} With comparable computational resources for unconditional image generation, we demonstrate that our multi-stage approach improves the Fréchet Inception Distance (FID) score for all SOTA methods. For example, on CIFAR-10 dataset \cite{krizhevsky2009learning}, our method improves the FID for DPM-Solver \cite{lu2022dpm} from 2.84 to \textbf{2.37}, and it improves the FID for EDM \cite{karras2022elucidating} from 2.05 (our training result) to \textbf{1.96}. Moreover, on the CelebA dataset \cite{liu2015faceattributes}, our approach significantly reduces the training computation of both EDM to \textbf{82\%} and the Latent Diffusion Model (LDM) \cite{rombach2022high} to \textbf{30\%}  for obtaining a similar quality of generation.
\end{itemize}

\paragraph{Organization.} 
In \Cref{sec:related_work}, we provide an overview of relevant literature to distinguish us from previous works. 
In \Cref{sec:multistage_framework}, we describe our proposed multistage framework for diffusion modeling, delineating the two core components. 
In \Cref{sec:analysis}, we present our observations and analysis that motivated the proposed multistage framework, justifying its development. 
Finally, in \Cref{sec:experiments}, we provide the results from our numerical experiments that validate the effectiveness of the proposed multistage approach.

%% file: section/relatedwork.tex
\section{Preliminaries \& Related Work}
\label{sec:related_work}

In this section, we start by reviewing the basic fundamentals of diffusion models \cite{ho2020denoising, song2020score, karras2022elucidating}. Subsequently, we delve into prior endeavors aimed at improving the training and efficiency of diffusion models through the partitioning of the timestep interval. Lastly, we revisit prior studies that significantly decrease the number of required sampling iterations.

\subsection{Background on Diffusion Models}
\label{sec:prelimilary}

Let $\bm{x}_0 \in \mathbb R^n$ denote a sample from the data distribution $p_{\text{data}}(\bm{x})$. Diffusion models operate within forward and reverse processes. The forward process progressively perturbs data $\bm{x}_0$ to a noisy version $\bm{x}_{t \in [0, 1]}$ via corrupting with the Gaussian kernel. This process can be formulated as a stochastic differential equation (SDE) ~\cite{song2020score} of the form 
$\text{d}\bm x = \bm x_t f(t)\text{d}t + g(t) \text{d} \bm{w}_t$, where $f(t)$ and $g(t)$ are the drift and diffusion coefficients, respectively, that correspond to a pre-defined noise schedule. $\bm{w}_t \in \mathbb R^n$ is the standard Wiener process. Under the forward SDE, the perturbation kernel is given by the conditional distribution defined as
\begin{equation}
    p_{t}(\bm x_t|\bm x_0) = \mathcal{N}(\bm x_t;s_t \bm x_0, s_t^2\sigma_t^2\textbf{I})\:,
\label{eq:pertb}
\end{equation}
where
\begin{equation}\label{eq:noise_parameters}
    s_t = \text{exp}(\int_0^t f(\xi)\text{d}\xi), \ \ \text{and}\  \sigma_t = \sqrt{\int_{0}^{t}\frac{g^2(\xi)}{s_{\xi}^2}\text{d}\xi}\:. 
\end{equation}
The parameters defined in \cref{eq:noise_parameters} are designed such that: (i) the data distribution is approximately estimated when $t=0$, and (ii) a nearly standard Gaussian distribution is obtained when $t=1$.
The objective of diffusion models is to learn the corresponding reverse SDE, defined as
\begin{align}
\label{eq:reverse_sde}
    \text{d}\bm{x} = \left[  f(t)\bm{x}_t - g^2(t) \nabla_{\bm{x}_t} \log p_t(\bm{x}_t)\right]\text{d}t +g(t)\text{d}\bar{\bm{w}},
\end{align}
where $\bar{\bm{w}} \in \mathbb R^n$ is the standard Wiener process running backward in time, and $\nabla_{\bm{x}_t} \log p_t(\bm{x}_t)$ is the (Stein) score function. In practice, the score function is approximated using a neural network $\bm \eps_{\bm \theta}:\mathbb R^n \times [0,1] \rightarrow \mathbb R^n $ parameterized by $\bm \theta$, which can be trained by the denoising score matching technique \cite{vincent2011connection} as
%
\begin{equation}
    \label{eq:score_matching}
    \min\limits_{\bm \theta} \mathbb{E}\left[ \omega(t) \|  \bm{\epsilon}_{\bm{\theta}}(\bm{x}_t, t) + s_t \sigma_t \nabla_{\bm{x}_t} \log p_t(\bm{x}_t | \bm{x}_0)\|_2^2 \right],
\end{equation}
which can also be written as $\min_{\bm \theta} \mathbb{E} [ \omega(t) ||\bm{\epsilon}_{\bm{\theta}}(\bm x_t, t) - \bm{\epsilon}||^2] + C$, where the expectation is taken over $t \sim [0, 1] $, $\bm{x}_t \sim p_t(\bm{x}_t | \bm{x}_0)$, $\bm{x}_0 \sim p_{\text{data}}(\bm{x})$, and $\bm \epsilon \sim \mathcal N (\textbf 0, \textbf I)$. Here, $C$ is a constant independent of $\bm\theta$, and $\omega(t)$ is a scalar representing the weight of the loss as a function of $t$. In DDPM \cite{ho2020denoising}, it is simplified to $\omega(t)=1$. Once the parameterized score function $\bm \eps_{\bm \theta}$ is trained, it can be utilized to approximate the reverse-time SDE using numerical solvers such as Euler-Maruyama.

\subsection{Timestep Clustering}


Diffusion models have demonstrated exceptional performance but face efficiency challenges in training and sampling. In response, several studies proposed to cluster the timestep range $t \in [0, 1]$ into multiple intervals (e.g., $[0, t_1), [t_1, t_2), \ldots, [t_n, 1]$). Notably, Choi et al. \cite{choi2022perception} reconfigured the loss weights for different intervals to enhance performance. Balaji et al. \cite{balaji2022ediffi} introduced ``expert denoisers'', which proposed using distinct architectures for different time intervals in text-to-image diffusion models. Go et al. \cite{go2023towards} further improved the efficiency of these expert denoisers through parameter-efficient fine-tuning and data-free knowledge transfer. Lee et al. \cite{lee2023multi} designed separate architectures for each interval based on frequency characteristics. Moreover, Go et al. \cite{go2023addressing} treated different intervals as distinct tasks and employed multi-task learning strategies for diffusion model training, along with various timestep clustering methods.

Our approach distinguishes itself from the aforementioned methods in two key aspects. The first key component is our tailored U-net architecture using a unified encoder coupled with different decoders for different intervals, while previous models have either adopted a unified architecture, as seen in \cite{choi2022perception, go2023addressing}, or employed separate architectures (referred to as expert denoisers) for each interval \cite{balaji2022ediffi, go2023towards, lee2023multi}. In comparison, our multistage architecture surpasses these methodologies, as demonstrated in \Cref{sec:compare_architecture}. Second, we developed a new timestep clustering method leveraging a general optimal denoiser (Proposition \ref{prop:diffusion model optimal solution}) that showcases superior performance (see \Cref{sec:compare_clustering}). In contrast, prior works rely on (i) a simple timestep-based clustering cost function \cite{balaji2022ediffi, go2023addressing, go2023towards, lee2023multi}, (ii) Signal-to-Noise Ratio (SNR) based clustering \cite{go2023addressing}, or (iii) gradients-based partitioning that uses task affinity scores \cite{go2023addressing}.

\subsection{Reducing the Sampling Iterations}



Efforts to improve sampling efficiency of diffusion models have led to many recent advancements in SDE and Ordinary Differential Equation (ODE) samplers \cite{song2020score}. For instance, the Denoising Diffusion Implicit Model (DDIM) \cite{song2020denoising} formulates the forward diffusion as a non-Markovian process with a deterministic generative path, significantly reducing the number of function evaluations (NFE) required for sampling (from thousands to hundreds) without sacrificing generation quality. Generalized DDIM (gDDIM) \cite{zhang2022gddim} further optimized DDIM by modifying the parameterization of the scoring network, reducing NFE to nearly 27. Furthermore, the works in \cite{lu2022dpm} and \cite{zhang2022fast}, termed the Diffusion Probabilistic Model solver (DPM-solver) and the Diffusion Exponential Integrator Sampler (DEIS), respectively, introduced fast higher-order solvers, employing exponential integrators that require even less than 20 NFE for comparable generation quality. Moreover, the consistency model \cite{song2023consistency} introduced a novel training loss and parameterization, achieving high-quality generation with merely 1-2 NFE.

We remark that while the aforementioned methods are indirectly related to our work, our experiments in \Cref{sec:superior_generation} and \Cref{sec:efficiency} show that our approach can be easily integrated into these techniques, further improving diffusion models' overall training and sampling efficiency.

%% file: section/method.tex
\section{Proposed Multistage Framework}
\label{sec:multistage_framework}

In this section, we introduce our new multistage framework (as illustrated in \Cref{fig:arch} (c)). Specifically, we first introduce the multi-stage U-Net architecture design in \Cref{sec:architecture}, following a new clustering method for choosing the optimal interval check points to partition the entire timestep $[0, 1]$ into intervals in \Cref{sec:clustering_alg}. We leave the discussion on the motivation behind our method to \Cref{sec:analysis}.

\subsection{Proposed Multi-stage U-Net Architectures}
\label{sec:architecture}

Most existing diffusion models either employ a unified architecture across all intervals \cite{ho2020denoising, song2020score, karras2022elucidating} to share features for all timesteps, or they use completely separate architectures for different timestep intervals \cite{lee2023multi, go2023towards, balaji2022ediffi} 
to take advantage of benign properties within different intervals. The details of the advantages and disadvantages of each architecture are discussed in section \ref{sec:analysis}.

To harness the advantages of both unified and separate architectures employed in prior studies, we introduce a multistage U-Net architecture, as illustrated in Figure~\ref{fig:arch}(c). Specifically, we partition the entire timestep $[0, 1]$ into several intervals, e.g., three intervals $[0, t_1), [t_1, t_2), [t_2, 1]$ in  \Cref{fig:arch}. For the architecture, we introduce:
\begin{itemize}[leftmargin=*]
    \item \textbf{One shared encoder across all time intervals.} For each timestep interval, we implement a shared encoder architecture (plotted in blue in \Cref{fig:arch} (c)), which is similar to the architecture employed in the original U-Net framework \cite{ronneberger2015u}. Unlike separate architecture, the shared encoder provide shared information across all timesteps, preventing models from overfitting (see \Cref{sec:method_architecture} for a discussion).  
    \item \textbf{Separate decoders for different time intervals.}  Motivated by the multi-head structure introduced in the Mask Region-based Convolutional Neural Networks (Mask-RCNN) method \cite{he2017mask}, we propose to use multiple distinct decoders (plotted in colors for different intervals in \Cref{fig:arch}(c)), where each decoder is tailored to a specific timestep interval. The architecture of each decoder closely resembles the one utilized in \cite{song2020score}, with deliberate adjustments made to the embedding dimensions to optimize performance.
\end{itemize}
As we observe, the primary difference in the architecture resides within the decoder structure. Intuitively, we use a decoder with fewer number of parameters for intervals closer to the noise, because the learning task is easier. We use a decoder with larger number of parameters for intervals closer to the image.
In Section \ref{sec:analysis}, we conduct a comprehensive analysis to elucidate the rationale behind the adoption of this multistage architecture, and the necessity for varying parameters across different intervals.

\subsection{Optimal Denoiser-based Timestep Clustering}
\label{sec:clustering_alg}

Next, we discuss how we principally choose the interval partition time points in practice. For simplicity, we focus on the case where we partition the time $[0,1]$ into three intervals $[0, t_1), [t_1, t_2), [t_2, 1]$, and we develop a timestep clustering method to choose the optimal $t_1$ and $t_2$ that we introduce in the following. Of course, our method can be generalized to multi-stage networks with arbitrary interval numbers. However, in practice, we find that the choice of three intervals strikes a good balance between effectiveness and complexity; see our ablation study in the appendices.


To partition the time interval, we employ the optimal denoiser established in Proposition \ref{prop:diffusion model optimal solution}, where we can show the following result.

\begin{algorithm}[t]
\caption{Optimal Denoiser based Timestep Clustering}\label{alg:cap}

\begin{algorithmic}[1]
\State \textbf{Input}: Total samples $K$, optimal denoiser function $\bm{\epsilon}^*_{\bm \theta} (\bm x, t)$, thresholds $\alpha$, $\eta$, dataset $p_{\text{data}}$
\State \textbf{Output}: Timesteps $t_1$, $t_2$
\State $\mathcal{S}_0 = \mathcal{S}_1 = \emptyset$
\For{$k\in \{1,\dots,K\}$}
\State $\bm y_k \sim p_{\text{data}}, \bm{\epsilon}_k \sim \mathcal{N}(0, \mathbf{I}), t_k \sim [0, 1]$
\State $\mathcal{S}^k_0 \leftarrow \mathcal{D}(\bm{\epsilon}^*_{t_k},  \bm{\epsilon}^*_{0}, \bm y_k, \bm{\epsilon}_k), \mathcal{S}^k_1 \leftarrow \mathcal{D}(\bm{\epsilon}^*_{t_k},  \bm{\epsilon}^*_{1}, \bm y_k, \bm{\epsilon}_k)$
\State $\mathcal{S}_0 \leftarrow \mathcal{S}_0 \cup \{ \left(t_k,  \mathcal{S}^k_0 \right)\}$ 
\State $\mathcal{S}_1 \leftarrow \mathcal{S}_1 \cup \{ \left(t_k,  \mathcal{S}^k_1 \right)\}$
\EndFor
\State $t_1 =\argmax\limits_{\tau}  \left\{\tau~\Big|~\dfrac{\sum_{(t_k,\mathcal{S}^k_0) \in \mathcal{S}_0} \left[\mathcal{S}^k_0 \cdot \mathds{1}(t_k \leq \tau)\right]}{\sum_{(t_k,\mathcal{S}^k_0) \in \mathcal{S}_0} \left[\mathds{1}(t_k \leq \tau)\right]} \geq \alpha \right\}$
\State $t_2 = \argmin\limits_{\tau}  \left\{\tau~\Big|~\dfrac{\sum_{(t_k,\mathcal{S}^k_1) \in \mathcal{S}_0}\left[\mathcal{S}^k_1 \cdot \mathds{1}(t_k \geq \tau)\right]}{\sum_{(t_k,\mathcal{S}^k_1) \in \mathcal{S}_0} \left[\mathds{1}(t_k \geq \tau)\right]}\geq \alpha \right\}$
\end{algorithmic}
\label{alg:interval}
\end{algorithm}

\begin{prop}
\label{prop:diffusion model optimal solution}
    Suppose we train a diffusion model denoiser function $\bm \eps_{\bm \theta} (\bm x,t)$ with parameters $\bm \theta$ using dataset $\Brac{\bm y_i \in \mathbb{R}^n}_{i=1}^N $ by 
    \begin{align}
    \label{eqn:training-loss}
       \min_{\bm \theta}  \mathcal{L}(\bm{\epsilon}_{\bm{\theta}} ; t) = \mathbb{E}_{\bm x_0,\bm x_t, \epsilon} [||\bm{\epsilon} - \bm{\epsilon}_{\bm{\theta}}(\bm x_t, t)||^2],
    \end{align}
    where $\bm x_0 \sim p_{\text{data}}(\bm x) = \frac{1}{N} \sum_{i =1}^{N} \delta (\bm x - \bm y_i)$, $\bm \epsilon \sim \mathcal{N}(0,\textbf{I})$, and $ \bm x_t \sim  p_{t}(\bm x_t|\bm x_0) = \mathcal{N}(\bm x_t;s_t\bm x_0, s_t^2\sigma_t^2\textbf{I})$ with perturbation parameters $s_t$ and $ \sigma_t$ defined in \cref{eq:noise_parameters}. Then, the optimal denoiser at $t$, defined as $ \bm{\epsilon}^*_{\bm{\theta}}(\bm x;t) = \argmin_{\bm{\epsilon}_{\bm{\theta}}} \mathcal{L}(\bm{\epsilon}_{\bm{\theta}}; t)$, is given by 
    \begin{align*}
        \bm{\epsilon}^*_{\bm \theta}(\bm x;t) = \frac{1}{s_t\sigma_t}\brac{\bm x - s_t\frac{\sum_{i = 1}^{N}\mathcal{N}(\bm x;s_t\bm y_i, s_t^2\sigma_t^2\textbf{I})\bm y_i}{\sum_{i = 1}^{N}\mathcal{N}(\bm x;s_t\bm y_i, s_t^2\sigma_t^2\textbf{I})}}\:.
    \end{align*}
\end{prop}
The proof is provided in appendices, and the result can be generalized from recent work of Karras et al. \cite{karras2022elucidating}, extending from a specific kernel $p_{t}(\bm x_t|\bm x_0) = \mathcal{N}(\bm x_t; \bm x_0, \sigma_t^2\textbf{I})$ to encompassing a broader scope of noise perturbation kernels, given by $p_{t}(\bm x_t|\bm x_0) = \mathcal{N}(\bm x_t; s_t \bm x_0, s_t^2\sigma_t^2\textbf{I})$. For brevity, we simplify the notation of the optimal denoiser $\bm{\epsilon}^*_{\theta}(\bm x, t)$ in Proposition \ref{prop:diffusion model optimal solution} as $\bm{\epsilon}^*_{t}(\bm x)$.

To obtain the optimal interval, our rationale is to \emph{homogenize} the regression task as much as possible within each individual time interval.   
To achieve this goal, given sampled $\bm x_0$, $\bm{\epsilon}$, we define the function distance of the optimal denoiser at any given timestep $t_a$, $t_b$ as:
%
%
\begin{equation*}
    \mathcal{D}(\bm{\epsilon}^*_{t_a},  \bm{\epsilon}^*_{t_b}, \bm x_0, \bm{\epsilon}) = \frac{1}{n} \sum_{i=1}^{n} \mathds{1}(|\bm{\epsilon}^*_{t_a}(\bm x_{t_a}) - \bm{\epsilon}^*_{t_b}(\bm x_{t_b})|_{i} \leq \eta)\:,
\end{equation*}
%
where $\mathds{1}(\cdot)$ is the indicator function, $\eta$ is a pre-specified threshold, $ \bm x_{t_a} = s_{t_a} \bm x_0 + s_{t_a}\sigma_{t_a} \bm{\epsilon}$, and $\bm x_{t_b} = s_{t_b} \bm x_0 + s_{t_b}\sigma_{t_b} \bm{\epsilon}$. Consequently, we define the functional similarity of the optimal denoiser at timesteps $t_a$ and $t_b$ as:
\begin{equation}
      \mathcal{S}(\bm{\epsilon}^*_{t_a},  \bm{\epsilon}^*_{t_b}) = \mathbb{E}_{\bm x_0 \sim p_{\text{data}}} \mathbb{E}_{\bm{\epsilon} \sim \mathcal{N}(0, \textbf{I})}[\mathcal{D}(\bm{\epsilon}^*_{t_a},  \bm{\epsilon}^*_{t_b}, \bm x_0, \bm{\epsilon})]\:.
\end{equation}
Based upon the definition, we design the following optimization problem to find the largest $t_1$ and smallest $t_2$ as: 
\begin{align}
   t_1 &\leftarrow \argmax\limits_{\tau} \left\{\tau\Big|\mathbb{E}_{t \sim [0, \tau)} [\mathcal{S}(\bm{\epsilon}^*_t,  \bm{\epsilon}^*_0)] \geq \alpha \right\}, \\
   t_2 &\leftarrow \argmin\limits_{\tau} \left\{\tau\Big| \mathbb{E}_{t \sim [\tau, 1]} [\mathcal{S}(\bm{\epsilon}^*_t,  \bm{\epsilon}^*_1)] \geq \alpha \right\},
\end{align}
%
such that the average functional similarity of $\bm{\epsilon}^*_t$ (resp. $\bm{\epsilon}^*_t$) to $\bm{\epsilon}^*_0$ (resp. $\bm{\epsilon}^*_1$) in $[0, t_1)$ (resp. $[t_2, 1]$) is larger than or equal to a pre-defined threshold $\alpha$. As the above optimization problems are intractable, we propose the procedure outlined in \Cref{alg:interval} to obtain an approximate solution.\footnote{Basically, the algorithm samples $K$ sample pairs $(\bm y_k, \bm\epsilon_k, t_k), k\in\{1, \ldots, K\}$ to calculate the distance $\mathcal{D}(\bm{\epsilon}^*_{t_k},  \bm{\epsilon}^*_{0}, \bm y_k, \bm\epsilon_k)$ and $\mathcal{D}(\bm{\epsilon}^*_{t_k},  \bm{\epsilon}^*_{1}, \bm y_k, \bm\epsilon_k)$ and then based on those distances to solve the optimization problem define in the line 10 -11 of \Cref{alg:interval} to get $t_1$ and $t_2$.}

%% file: section/analysis.tex
\section{Identification of Key Sources of Inefficiency}
\label{sec:analysis}

Conventional diffusion model architectures, as exemplified by \cite{song2020score, ho2020denoising, karras2022elucidating}, treat the training of the diffusion model as a unified process across all timesteps. Recent research endeavors, such as \cite{choi2022perception, go2023addressing, lee2023multi, go2023towards}, have highlighted the benefits of recognizing distinctions between different timesteps and the potential efficiency gains from treating them as separate tasks during the training process. However, our experimental results demonstrate that both unified and separate architectures suffer inefficiency for training diffusion models, where the infficiency comes from (i) overfitting and (ii) gradient dissimilarity.



\subsection{Empirical Observations on the Key Sources of Inefficiency}

To illustrate the inefficiency in each interval, we isolate the interval by using a separate architecture from the rest. 


\begin{figure*}[t]
\centering
  \begin{subfigure}[t]{.45\linewidth}
    \centering\includegraphics[width=200pt,height=200pt]{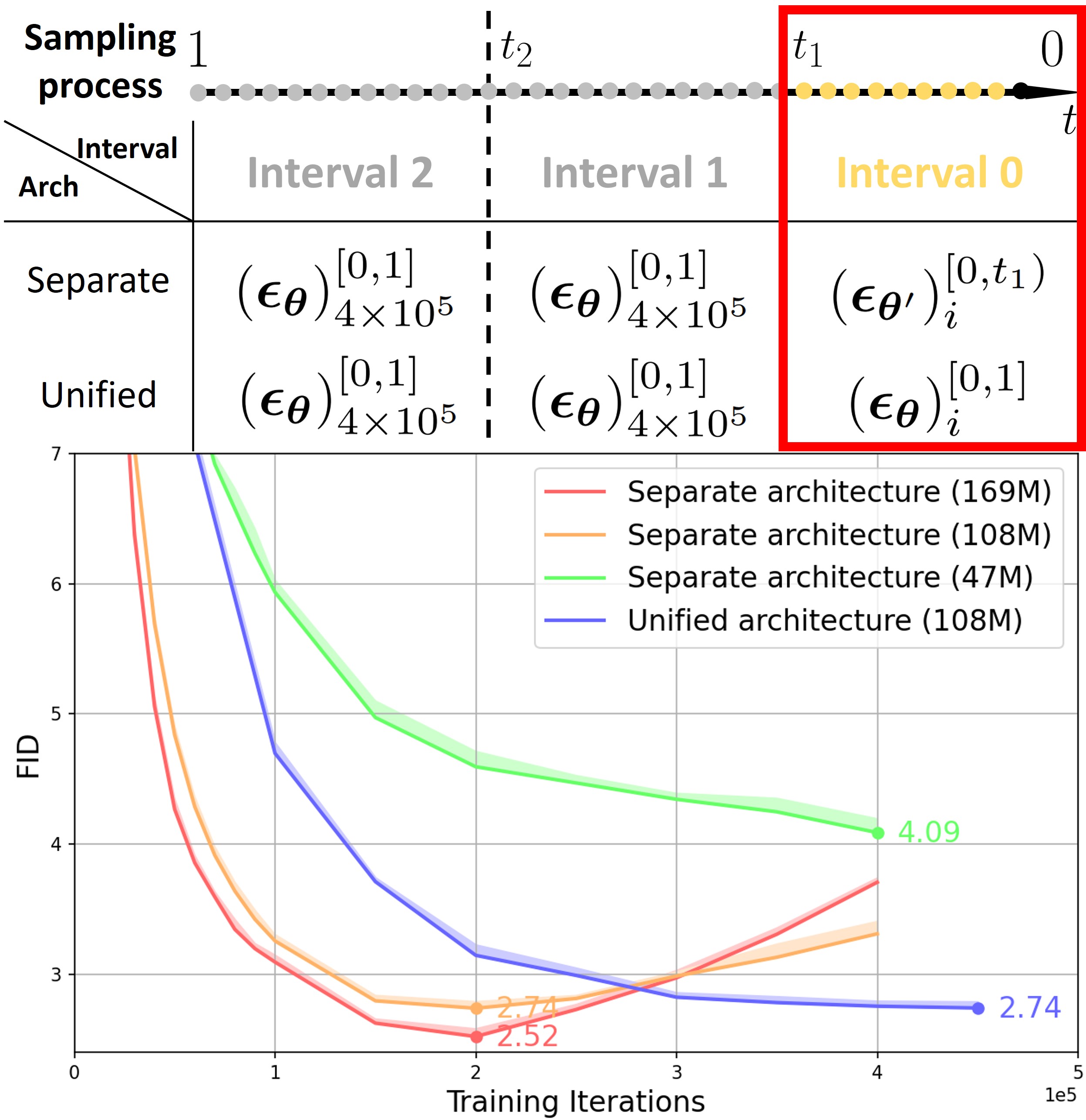}
    \caption{Analysis on interval $[0, t_1)$}
    \label{fig:analysis interval0}
  \end{subfigure}
\begin{subfigure}[t]{.45\linewidth}
    \centering\includegraphics[width=200pt,height=200pt]{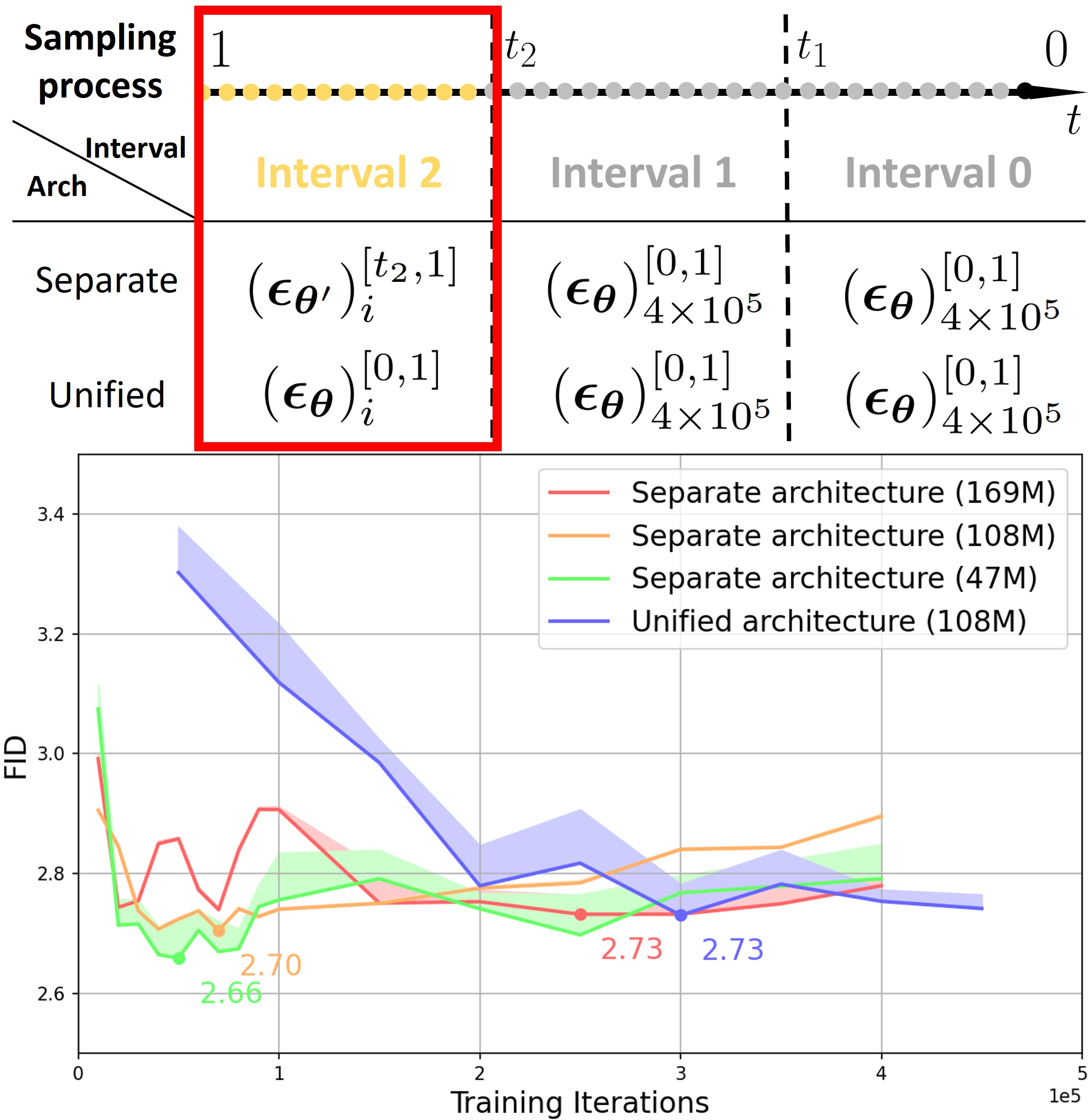}
    \caption{Analysis on interval $[t_2, 1]$}
    \label{fig:analysis interval2}
  \end{subfigure}

\caption{\textbf{Comparison between separate architecture and unified architecture w.r.t. the image generation quality in different intervals:} (a) analysis on Interval 
$[0, t_1)$; and (b) analysis on Interval  $[t_2, 1]$. As illustrated on top of each figure, we \emph{only} train separate architectures within specific intervals for the sampling process in both (a) and (b). For the remaining period of sampling, we use a well-trained diffusion model $\Scale[0.8]{(\bm{\epsilon}_{\bm\theta})^{[0,1]}_{4 \times 10^5}}$ to approximate the ground truth score function. As shown in the above figure of (a), e.g. for the separate architecture on interval 1, sampling utilizes trained model $\Scale[0.8]{(\bm{\epsilon}_{\bm\theta'})^{[0,t_1)}_{i}}$ for interval 0 and well-trained model $\Scale[0.8]{(\bm{\epsilon}_{\bm\theta})^{[0,1]}_{4 \times 10^5}}$ for interval 1 and 2. 
Notably, for both $\Scale[0.8]{(\bm{\epsilon}_{\bm\theta})^{[0, 1]}_{i}}$ and $\Scale[0.8]{(\bm{\epsilon}_{\bm\theta})^{[0, 1]}_{4\times10^5}}$, we are using the model with 108M parameters. For separate architecture, the number in the parentheses represents the number of parameters of the model $\Scale[0.8]{(\bm{\epsilon}_{\bm\theta'})^{[a, b]}_{i}}$. For example, for separate architecture (169M) in (a), the model $\Scale[0.8]{(\bm{\epsilon}_{\bm\theta'})^{[0, t_1)}_{i}}$ has 169M parameters for $\bm\theta'$. The bottom figures in (a-b) illustrate the FID of the generation from each architecture under different training iterations.
}
\label{fig:analysis}
\end{figure*}

\paragraph{Experiment setup.} In our experiments, we consider three-stage training and divide the time steps into three intervals: $[0, t_1), [t_1, t_2), [t_2, 1]$.\footnote{Details for interval clustering can be found in \Cref{sec:exp_setting_timestep_clustering}.} Let $(\bm{\epsilon}_{\bm\theta})^{[a, b]}_i, 0 \leq a < b \leq 1$ denote a U-Net architecture with parameter $\bm\theta$ trained with $i$ iterations and fed with data pairs $(\bm x_t, t)$, where $t \in [a, b]$. We then trained models using two different strategies: a unified architecture with 108M network parameters for all intervals, i.e., $(\bm{\epsilon}_{\bm\theta})^{[0, 1]}_i$, and separate architectures with varying network parameters (e.g., 47M, 108M, 169M) for each interval; e.g., $(\bm{\epsilon}_{\bm\theta})^{[0, t_1)}_i$ for the interval $[0, t_1)$, etc. 
It is worth noting that, apart from the differences in network parameters, we utilize the same network architecture (e.g., U-Net) for both the unified and the separate approaches. We assessed the training progress of each model by evaluating image generation quality at different training iterations. Notably, because some of the models are only trained on one interval, we need to provide a ground truth score for the other intervals. In \Cref{fig:analysis}, the sampling process is shown above and the experimental results are shown below.

\paragraph{Inefficiency in unified architectures.} From \Cref{fig:analysis}, we observe the following:
\begin{itemize}[leftmargin=*]
  \item \textit{Overparameterization and underfitting emerge simultaneously for unified architectures}. In \Cref{fig:analysis interval0}, we observe that increasing the number of parameters in Interval 0 can improve the image generation quality (as indicated by a lower FID score). In contrast, \Cref{fig:analysis interval2} reveals that increasing the number of parameters in Interval 2 has minimal impact on the quality of image generation. This implies that using a unified architecture will result in underfitting in Interval 0 and  overparameterization in Interval 2. The current unified architecture's parameter redundancy leaves significant room for improving its efficiency. To optimize the computational usage, we should allocate more parameters to Interval 0 while allocate fewer parameters to Interval 2.

  \item \textit{Gradient dissimilarity hinders training for unified architecture.}
    Quantitative results from \cite{go2023addressing} demonstrate dissimilarity in gradients emerges among different intervals. This can also be observed from our results based upon both \Cref{fig:analysis interval0} and \Cref{fig:analysis interval2}. 
    For the unified and separate architectures using the same number of parameters (108M), separate architecture achieves a significantly lower FID with the same training iterations, implying that dissimilar gradients among intervals may hinder training when using a unified architecture. Here, the only difference between training separate and unified architectures is that the batch gradient for unified architecture is calculated based on all timesteps while the batch gradient for separate architecture is calculated only from a specific interval.
\end{itemize}


\paragraph{Inefficiency in existing separate architectures.}
Although separated architecture \cite{choi2022perception, lee2023multi, go2023towards} better allocates computational resources for each interval, it suffers from overfitting. This can be illustrated based upon training separate architectures (169M) and (108M) in Interval 0 shown in \Cref{fig:analysis interval0}, where increasing the number of parameters will lead to overfitting when we increase the number of training iterations beyond achieving the best FID. This also happens in Interval 2, when we compare all separate architectures in \Cref{fig:analysis interval2}. In comparison, the unified networks with 108M parameters are less prone to overfit for both Interval 0 and Interval 2. This suggests that we can reduce overfitting by training shared weights across different intervals together.


\subsection{Tackling the Inefficiency via Multistage U-Net Architectures}
\label{sec:method_architecture}

In a unified architecture applied across all timesteps, there is often a dual challenge: requirements for more parameters (169M) in the interval $[0, t_1)$ but fewer parameters (47M) in the interval $[t_2, 1]$. This issue is compounded by the gradient dissimilarity across different timesteps, which can impede effective training. Alternatively, employing separate architectures for different intervals might lead to overfitting and a lack of robust early stopping mechanisms. To address these challenges, our proposed multistage architecture in \Cref{sec:multistage_framework} combines shared parameters to reduce overfitting with interval-specific parameters to mitigate the impact of gradient dissimilarity. This tailored approach for each interval ensures improved adaptability. Furthermore, we conduct an in-depth ablation study in \Cref{sec:compare_architecture} to showcase the effectiveness of our multi-stage architecture over the existing models.

%% file: section/experiments_arxiv.tex
\section{Experiments}\label{sec:experiments}

For this section, we start with providing the basic experimental setups. Followed by this, in \Cref{sec:superior_generation}, we demonstrate the improved image generation quality of our multistage architecture over the state-of-the art models, under comparable training and sampling computations. On the other hand, in \Cref{sec:efficiency}, we demonstrate the improved training and sampling efficiency of our methods over existing methods, when compared with similar image generation quality. Finally, in \Cref{sec:compare_architecture} and \Cref{sec:compare_clustering},
we conduct comprehensive ablation studies on timestep clustering methods and multistage architectures, illustrating the superiority of the choices in our method.

\vspace{-0.1in}
\paragraph{Settings of the timestep clustering method.} \label{sec:exp_setting_timestep_clustering} We implement a Variance Preserving (VP) \cite{ho2020denoising, karras2022elucidating} perturbation kernel for clustering the time interval. In particular, we use $s_t = \sqrt{e^{\frac{1}{2}\beta_d t^2 + \beta_{\text{min}} - 1}}, \sigma_t = 1/\sqrt{e^{\frac{1}{2}\beta_d t^2 + \beta_{\text{min}}}}, t \in [\epsilon_t,1], \beta_d = 19.9, \beta_{\text{min}} = 0.1$, and $, \epsilon_t = 10^{-3}$. To obtain $t_1$ and $t_2$, we utilize the CIFAR-10 \cite{krizhevsky2009learning} with dataset size $N = 5\times10^4$. Specifically, Algorithm~\ref{alg:interval} is configured with $\eta = \frac{2}{256}, \alpha = 0.9,$ and $ K = 5\times10^4$. This choice of $\eta$ ensures the similarity of measurements in the RGB space. We divide $t\in [0,1]$ into $10^{4}$ discrete time steps. Consequently, we employ a grid search to determine optimal values for $t_1$ and $t_2$ using the procedure outlined in Algorithm~1. For the CelebA dataset \cite{liu2015faceattributes}, we utilize the same $t_1$ and $t_2$. And for Variance Exploding (VE) perturbation kernel, we calculate an equivalent $\sigma_1$ and $ \sigma_2$.\footnote{VE utilizes $\sigma$ instead of $t$ to represent the timestep. We choose $\sigma_1$ and $ \sigma_2$ such that $\text{SNR}_{\text{VE}}(\sigma_1) =\text{SNR}_{\text{VP}}(t_1)$ and $ \text{SNR}_{\text{VE}}(\sigma_2) =\text{SNR}_{\text{VP}}(t_2)$. 
Further details are provided in the appendices.
} Results of Section \ref{sec:compare_clustering} verify that these choices of $t_1$ and $t_2$ return the best performance when compared to other clustering method.


\paragraph{Multistage architectures.} \label{sec:exp_setting_architection} 
Our multistage architecture, inspired by the U-Net model \cite{ronneberger2015u} used in DDPM++ \cite{ho2020denoising, karras2022elucidating, song2020score}, is modified for interval-specific channel dimensions. The proposed architecture is adopted to three diffusion models: DPM-Solver \cite{lu2022dpm}, Elucidating Diffusion Model (EDM) \cite{karras2022elucidating}, and Latent Diffusion Model (LDM) \cite{rombach2022high}. In particular, for the cases of DPM-Solver and EDM, the encoder's channel dimensions are standardized at 128, while the decoders are configured with 192, 128, and 16 channels for intervals $[0, t_1)$, $[t_1, t_2)$, and $[t_2, 1]$, respectively. In the LDM case, we use 224 channels across the encoder for all intervals whereas the decoders are configured with 256, 192, and 128 channels for the respective intervals. To decide the specific number of parameters for each decoder, we apply ablation studies in the appendices. 


\paragraph{Datasets, Evaluation Metrics, \& Baselines.} \label{sec:exp_setting_evaluation} 
We use CIFAR-10 (32 $\times$ 32), CelebA (32 $\times$ 32), and CelebA (256 $\times$ 256) datasets for our experiments. To evaluate the performance of our multistage diffusion model in terms of the generation quality, we use the standard Fréchet Inception Distance (FID) metric \cite{heusel2017gans}, and the required Number of Function Evaluations (NFEs). We assess the sampling efficiency using giga-floating point operations (GFLOPs) per function evaluation. For both separate architecture and our multistage architecture, equivalent GFLOPs are computed as a weighted summation (based on the ODE Sampler steps within each interval) of GFLOPs for each interval. Training efficiency is evaluated using total training iterations multiplied by the floating point operations per function evaluation,\footnote{Here we simplify it by ignoring the FLOPs for backward propagation, which is approximately proportional to FLOPs of forward evaluation.} measured by peta-floating point operations (PFLOPs). We choose DDPM \cite{ho2020denoising}, Score SDE \cite{song2020score}, Poisson Flow Generative Models (PFGM) \cite{xu2022poisson}, DDIM \cite{song2020denoising}, gDDIM \cite{zhang2022gddim}, DEIS \cite{zhang2022fast}, DPM-solver \cite{lu2022dpm}, and EDM \cite{karras2022elucidating} as the baseline.

\subsection{Comparison of Image Generation Quality} 
\label{sec:superior_generation}
In this subsection, we demonstrate the effectiveness of our approach by comparing the image generation quality (measured by FID) with comparable training and sampling computations (measured by NFE). Specifically, \Cref{tab:quality} presents FID scores to measure the sampling quality, and NFEs to measure the number of sampling iterations required using the CIFAR-10 dataset. Our method is compared to 8 baselines. As observed, our multistage DPM-Solver outperforms DPM-Solver in terms of the reported FID values while requiring similar training iterations (both are $4.5 \times 10^5$), and model GFLOPs (18.65 for multistage DPM-Solver versus 17.65 for DPM-Solver). A similar observation holds when we compare our multistage EDM and the vanilla EDM, where we reduce FID from  2.05 to \textbf{1.96} by using the multi-stage architecture. Remarkably, utilizing only 20 NFE, our Multistage DPM-Solver returns the same FID score as the one reported for the PFGM method, which requires 147 NFEs. These results also highlight the adaptability of our framework to higher-order ODE solvers; see the 8th and last row of \Cref{tab:quality}. 



\begin{table}[t]
\centering

\begin{tabular}{lcc}
\toprule
METHOD & NFE ($\downarrow$) & FID ($\downarrow$) \\ 
\midrule
DDPM      & 1000    & 3.17    \\
Score SDE      &  2000   & 2.20    \\
PFGM      &  147   & 2.35    \\
DDIM      & 100    & 4.16    \\ 
gDDIM      & 20    & 2.97    \\ 
DEIS      & 20    & 2.86    \\
DPM-solver      & 20    & 2.73    \\
\midrule
\rowcolor{Gray}
\textbf{Multistage DPM-solver (Ours)}      & 20    & \textbf{2.35}    \\
&&\\
EDM      & 35    & 2.05    \\
\midrule
\rowcolor{Gray}
\textbf{Multistage EDM (Ours)}      & 35   & \textbf{1.96}    \\
\bottomrule
\end{tabular}
\caption{\textbf{Sampling quality on CIFAR-10 dataset.}}
\label{tab:quality}
\end{table}

\subsection{Comparison of Training \& Sampling Efficiency}
\label{sec:efficiency}


\begin{table}[t]
\centering
\resizebox{\columnwidth}{!}{%
\begin{tabular}{lcccc}
\toprule
Dataset/Method & Training Iterations ($\downarrow$) & GFLOPs ($\downarrow$) & Total Training PFLOPs ($\downarrow$) & FID ($\downarrow$) \\ 
\toprule
\textbf{CIFAR-10} $32 \times 32$ & & & & \\
\midrule
DPM-Solver \cite{karras2022elucidating}   & $4.5 \times 10^5$ & \textbf{17.65} & 7.94 & 2.73    \\
\rowcolor{Gray}
Multistage DPM-Solver (Ours)  & $\textbf{2.5} \times \textbf{10}^5 \textbf{(56\%)}$   & 18.65 (106\%) & \textbf{4.66 (59\%)} & 2.71 \\
\toprule
\textbf{CelebA} $32 \times 32$ & & & &\\
\midrule
EDM \cite{karras2022elucidating}   & $5.7 \times 10^5$ & \textbf{17.65} & 10.06 & 1.55    \\
\rowcolor{Gray}
Multistage EDM (Ours)  & $\textbf{4.3} \times \textbf{10}^5 \textbf{(75\%)}$   & 19.25 (109\%) & \textbf{8.28 (82\%)} & 1.44  \\
\toprule
\textbf{CelebA} $256 \times 256$ & & & &\\
\midrule
LDM \cite{rombach2022high}  &  $4.9\times 10^5$  & 88.39 & 43.31 & 8.29    \\
\rowcolor{Gray}
Multistage LDM (Ours)  & $\textbf{1.7} \times \textbf{10}^5 \textbf{(35\%)}$  & \textbf{76.19 (86\%)} & \textbf{12.95 (30.0\%)} & 8.38 \\
\bottomrule
\end{tabular}
}
\caption{\textbf{Training and sampling efficiency on more datasets.}}
\label{tab:efficiency}
\end{table}
In this subsection, we further demonstrate the superiority of our method by comparing the training and sampling efficiency under comparable image generation quality.
Specifically, in \Cref{tab:efficiency}, we present the number of training iterations, GFLOPs, and total training PFLOPs of our approach, DPM-solver, EDM, and LDM using CIFAR-10 and CelebA datasets. Using the CIFAR-10 dataset, our multistage DPM-solver achieves similar FID scores (2.71 vs 2.73) while requiring nearly half the training iterations when compared to the vanilla DPM solver. For the case of EDM (resp. LDM), our approach returns an FID score of 1.44 (resp. 8.29), requiring $1.4 \times 10^5$ ($3.2\times 10^5$) less iterations when compared to vanilla DPM (resp. LDM). For the cases of DPM and EDM, we can achieve substantial reduction of training iterations, which is demonstrated by a marginal increase in the number of GFLOPs (row 3 vs. row 4 and row 5 vs. row 6). For the LDM case,we also achieve a significant reduction of training iterations, demonstrated by a reduced number of GFLOPs (row 8 vs. row 9). These promising results highlight the significantly improved computational efficiency achieved by using the proposed multistage framework.  



\subsection{Comparison of Different Architectures}
\label{sec:compare_architecture}

In \Cref{sec:analysis}, we highlighted the limitations of both unified and separate diffusion model architectures in terms of training efficiency (see \Cref{fig:analysis}). In this part, we further illustrate these limitations through extensive experiments as shown in \Cref{tab:better_architecture}. Here, we use the U-Net architecture, trained on the CIFAR-10 dataset, and utilize the DPM-Solver for sampling. For the unified case, we use a single U-Net model with 128 channels. For the separate case, three distinct U-Nets are used by which, for each interval, we use a 128-channel decoder. For improved performance of the separate architecture, we implement two techniques: early stopping (ES) and tailored parameters (TP) to tackle the overfitting and parameter inefficiency discussed in \Cref{sec:analysis}. Under ES, the criteria is to stop training prior to overfitting. For the tailored parameters, the three U-Nets are configured with 192, 128, and 16 channels decoders for Intervals 0, 1, and 2, respectively.

\begin{wraptable}{r}{0.6\linewidth}
\centering
\begin{tabular}{lcc}
\toprule
Method & GFLOPs & FID ($\downarrow$) \\ \midrule
Unified      & 17.65  & 2.73    \\
Separate       & 17.65  &  2.87   \\
Separate  (+ ES)      & 17.65  &  2.80   \\
Separate  (+ TP)      & 18.65  &  2.68   \\
Separate  (+ ES, TP)      & 18.65  & 2.52    \\
\midrule
\rowcolor{Gray}
\textbf{Multistage (Ours)}      & 18.65  & \textbf{2.35}    \\
\bottomrule
\end{tabular}
\caption{\textbf{Sampling quality on CIFAR-10 dataset for different diffusion model architectures.} Here, ES denotes early stopping, and TP denotes tailored parameters.}
\label{tab:better_architecture}
\end{wraptable}
 Our comparison and analysis in \Cref{tab:better_architecture} reveals notable insights of our network design. Comparisons between the 2nd and 3rd rows (and between the 4th and 5th rows) on the separate architectures indicate that early stopping effectively mitigates overfitting and enhances generation quality. When comparing the 2nd and 4th rows (as well as the 3rd and 5th rows) on the separate architectures, we observe that optimizing parameter usage can achieve a significant decrease in FID under comparable GFLOPs (18.65 vs. 17.65). Most importantly, our multistage architecture, as shown in the 6th row, benefits from both unified and separate architectures, achieving the best FID (2.35, compared to 2.73 and 2.52). Comparing the 2nd row and 4th row, the shared encoder not only prevents overfitting but also improves the convergence of the diffusion model as reported by the FID scores.

\subsection{Comparison of Timestep Clustering Methods}
\label{sec:compare_clustering} 


\begin{table}[t]
\centering
\begin{tabular}{lccc}
\toprule
Clustering Method & $t_1$ & $t_2$ & FID ($\downarrow$) \\ \midrule
Timestep \cite{go2023addressing,lee2023multi}     & 0.330 & 0.670   & 3.12    \\
SNR \cite{go2023addressing}    & 0.348 & 0.709   & 2.72    \\
Gradient \cite{go2023addressing}    & 0.360 & 0.660   & 2.75    \\
\midrule
\rowcolor{Gray}
\textbf{Optimal Denoiser (Ours)}      & 0.442 & 0.631  & \textbf{2.35}    \\
\bottomrule
\end{tabular}
\caption{\textbf{Sampling quality on CIFAR-10 dataset for different clustering methods.}}
\label{tab:cluster}
\end{table}
As preciously stated in \Cref{sec:related_work}, various timestep clustering methods are proposed including timestep-based, SNR-based, and gradient-based clustering approaches \cite{go2023addressing, lee2023multi}. In this subsection, we conduct an experiment to demonstrate the superiority of our clustering method compared to previous arts. Specifically, we apply the clustering methods in \cite{go2023addressing, lee2023multi} to partition the interval along with our proposed multistage UNet architecture. The computed intervals are shown in the Table~\ref{tab:cluster}. We use the multistage DPM-Solver with these different intervals trained on the CIFAR-10 dataset. As observed, our optimal denoiser-based clustering method achieves the highest FID score, consistently outperforming all other clustering methods.

%% file: section/conclusion.tex
\section{Conclusion}



In this study, we significantly enhance the training and sampling efficiency of diffusion models through a novel multi-stage framework. This method divides the time interval into several stages, using a specialized multi-decoder U-net architecture that combines time-specific models with a common encoder for all stages. Based on our practical findings, this multi-stage approach utilizes unique parameters for each timestep, along with shared parameters across all timesteps, grouped according to the most effective denoiser. We conducted thorough numerical experiments with three leading-edge diffusion models, such as large-scale latent diffusion models, which confirmed the effectiveness of our strategy.

In future research, it would be interesting to expand our multi-stage approach beyond the unconditional diffusion models, such as conditional diffusion models and diffusion models for solving inverse problems. Our experiments, as detailed in \Cref{sec:efficiency}, demonstrate that training latent diffusion models within our multi-stage framework requires only 30\% of the computational effort needed for training standard latent diffusion models on the CelebA dataset. Thus, employing a multi-stage strategy could significantly reduce the computational demands for training large-scale stable diffusion models, such as those described in \cite{rombach2022high}, which typically requires significant computational power.

%% file: section/appendix_theorem_arxiv.tex
We include proof of Proposition~1 in \Cref{appendix:proof}, extension of Algorithm \Cref{alg:interval} for more intervals in \Cref{append:more interval}, provide generation visualization in \Cref{append:visualization}, conduct ablation study on network parameters in \Cref{append:abla_num_parameter} and number of intervals in \Cref{append:abla_num_interval}, and describe experimental settings for different interval splitting from \Cref{sec:compare_clustering} in \Cref{append:exp_set_interval_split}.

\section{Proof of Proposition~1}
\label{appendix:proof}

\subsection{Background}

This section presents the proof of Proposition~\ref{prop:diffusion model optimal solution} in \Cref{sec:multistage_framework}. Our proof partially follows \cite{karras2022elucidating}. As background, let $p_{t}(\bm x_t|\bm x_0) = \mathcal{N}(\bm x_t;s_t \bm x_0, s_t^2\sigma_t^2\textbf{I})$ be the perturbation kernel of the diffusion model, which is a continuous process gradually adding noise from original image $\bm x_0$ to $\bm x_t$ across $t \in [0, 1]$. Both $s_t = s(t)$ and $ \sigma_t = \sigma(t)$ are used interchangeably to denote scalar functions of $t$ that control the perturbation kernel. It has been shown in \cite{song2020score} that this perturbation kernel is equivalent to a stochastic differential equation $\text{d}\bm x = \bm x_t f(t)  \text{d}t + g(t) \text{d} \bm \omega_t$, where $f(t), g(t)$ are scalar functions of $t$. The relations of $f(t), g(t)$ and $s_t, \sigma_t$ are: 
%
\begin{equation}
    s_t = \text{exp}(\int_0^tf(\xi)\text{d}\xi), \ \ \text{and}\  \sigma_t = \sqrt{\int_{0}^{t}\frac{g^2(\xi)}{s^2(\xi)}\text{d}\xi}\:.
\end{equation}
%
\subsection{Proof}
Given dataset $\Brac{\bm y_i }_{i=1}^N$ with $N$ images, we approximate the original dataset distribution as $p_{\text{data}}$ as multi-Dirac distribution, $p_{\text{data}}(\bm x) = \frac{1}{N}\sum_{i = 1}^{N} \delta (\bm x - \bm y_i)$. Then, the distribution of the perturbed image $\bm x$ at random timestep $t$ can be calculated as:
\begin{align}
    p_t(\bm x) 
    &= \int_{\mathbb{R}^d} p_{t}(\bm x|\bm x_0) p_{\text{data}}(\bm x_0) \text{d}\bm x_0 \\
    &= \int_{\mathbb{R}^d} p_{\text{data}}(\bm x_0) \mathcal{N}(\bm x;s_t\bm x_0, s_t^2\sigma_t^2\textbf{I}) \text{d}\bm x_0 \\
    &= \int_{\mathbb{R}^d} \frac{1}{N}\sum_{i = 1}^{N} \delta (\bm x_0 - \bm y_i) \mathcal{N}(\bm x;s_t\bm x_0, s_t^2\sigma_t^2\textbf{I}) \text{d}\bm x_0 \\
    &= \frac{1}{N}\sum_{i = 1}^{N} \int_{\mathbb{R}^d} \delta (\bm x_0 - \bm y_i) \mathcal{N}(\bm x;s_t\bm x_0, s_t^2\sigma_t^2\textbf{I}) \text{d}\bm x_0 \\
    &= \frac{1}{N}\sum_{i = 1}^{N} \mathcal{N}(\bm x;s_t\bm y_i, s_t^2\sigma_t^2\textbf{I})\:.
\end{align}
%
Let the noise prediction loss, which is generally used across various diffusion models, be
\begin{align}
    \mathcal{L}(\bm \epsilon_{\bm\theta}; t) 
    &= \mathbb{E}_{\bm x \sim p_t(\bm x)} [|\bm \epsilon - \bm \epsilon_{\bm \theta}(\bm x, t)||^2] = \int_{\mathbb{R}_d} \frac{1}{N}\sum_{i = 1}^{N} \mathcal{N}(\bm x;s_t\bm y_i, s_t^2\sigma_t^2\textbf{I}) ||\bm \epsilon - \bm\epsilon_{\bm\theta}(\bm x, t)||^2 \text{d}\bm x \:,
    \label{eq:loss}
\end{align}
where $\bm \epsilon \sim \mathcal{N}(\bm 0, \textbf{I})$ is defined follow the perturbation kernel $p_{t}(\bm x|\bm x_0) = \mathcal{N}(\bm x;s_t \bm x_0, s_t^2\sigma_t^2\textbf{I})$, and
\begin{equation}
   \bm x = s_t\bm y_i + s_t \sigma_t \bm \epsilon \quad \Rightarrow \quad \bm \epsilon = \frac{\bm x - s_t\bm y_i}{s_t \sigma_t}\:.
\label{eq:eps_x_relation}
\end{equation}
Here, $\bm \epsilon_{\bm \theta}$ is a "denoiser" network for learning the noise $\bm \epsilon$. Plugging \eqref{eq:eps_x_relation} into \eqref{eq:loss}, the loss can be reparameterized as:
%
%
\begin{equation}
    \mathcal{L}(\bm \epsilon_{\bm\theta}; t) = \int_{\mathbb{R}_d} \mathcal{L}(\bm \epsilon_{\bm\theta};\bm x, t) \text{d}\bm x \:,
\label{eq:loss_x}
\end{equation}
where
\begin{align*}
    &\mathcal{L}(\bm \epsilon_{\bm\theta};\bm x, t) = \frac{1}{N}\sum_{i = 1}^{N} \mathcal{N}(\bm x;s_t\bm y_i, s_t^2\sigma_t^2\textbf{I}) ||\bm \epsilon_{\bm\theta}(\bm x, t) - \frac{\bm x - s_t\bm y_i}{s_t \sigma_t}||^2\:.
\end{align*}
Eq.\eqref{eq:loss_x} means that we can minimize $\mathcal{L}(\bm \epsilon_{\bm\theta}; t)$ by minimizing $\mathcal{L}(\bm \epsilon_{\bm\theta};\bm x, t)$ for each $\bm x$. As such, we find the "optimal denoiser" $\bm \epsilon^*_{\bm\theta}$ that minimize the $\mathcal{L}(\bm \epsilon_{\bm\theta};\bm x, t)$, for every given $\bm x$ and $t$, as:
\begin{equation}
    \bm \epsilon^*_{\bm\theta}(\bm x;t) = \text{arg} \ \text{min}_{\bm \epsilon_{\bm\theta}(\bm x;t)} \mathcal{L}(\bm \epsilon_{\bm\theta};\bm x, t)\:.
\end{equation}
The above equation is a convex optimization problem by which the solution can be obtained by setting the gradient of $\mathcal{L}(\bm \epsilon_{\bm\theta};\bm x, t)$ w.r.t $\bm \epsilon_{\bm\theta}(\bm x;t)$ to zero. This means
\begin{align}
    &\nabla_{\bm \epsilon_{\bm\theta}(\bm x;t)} [\mathcal{L}(\bm \epsilon_{\bm\theta};\bm x, t)] = 0 \\
    \Rightarrow & \frac{1}{N}\sum_{i = 1}^{N} \mathcal{N}(\bm x;s_t\bm y_i, s_t^2\sigma_t^2\textbf{I}) [\bm\epsilon^*_{\bm\theta}(\bm x;t) - \frac{\bm x - s_t\bm y_i}{s_t \sigma_t}] = 0 \\
    \Rightarrow & \bm\epsilon^*_{\bm\theta}(\bm x;t) = \frac{1}{s_t\sigma_t}[\bm x - s_t\frac{\sum_{i = 1}^{N}\mathcal{N}(\bm x;s_t\bm y_i, s_t^2\sigma_t^2\textbf{I})\bm y_i}{\sum_{i = 1}^{N}\mathcal{N}(\bm x;s_t\bm y_i, s_t^2\sigma_t^2\textbf{I})}]
    \label{eq:optim_func}
\end{align}

%% file: section/appendix_threestage_split_arxiv.tex
\section{Generalized \Cref{alg:interval}}

\label{append:more interval}

\begin{algorithm*}[t]
\caption{Optimal Denoiser based Timestep Clustering for More Intervals}

\begin{algorithmic}[1]
\State \textbf{Input}: Total samples $K$, optimal denoiser function $\bm{\epsilon}^*_{\bm \theta} (\bm x, t)$, interval number $n$, dataset $p_{\text{data}}$
\State \textbf{Output}: Timesteps $t_1$, $t_2$, $\ldots$, $t_{n-1}$
\State $\mathcal{S} \leftarrow \emptyset$
\For{$k\in \{1,\dots,K\}$}
\State $\bm y_k \sim p_{\text{data}}, \bm{\epsilon}_k \sim \mathcal{N}(0, \mathbf{I}), t_{a_k} \sim [0, 1] , t_{b_k} \sim [0, 1]$
\State $\mathcal{S}^k \leftarrow \mathcal{D}(\bm{\epsilon}^*_{t_{a_k}},  \bm{\epsilon}^*_{t_{b_k}}, \bm y_k, \bm{\epsilon}_k)$
\State $\mathcal{S} \leftarrow \mathcal{S} \cup \{ \left(t_{a_k}, t_{b_k}, \mathcal{S}^k \right)\}$ 
\EndFor
\State $t_1, \cdots, t_{i-1} \leftarrow \argmin_{t_1, \cdots, t_{i-1}} \quad \sum\limits_{k = 1}^{K}\sum\limits_{i = 0}^{n - 1}\sum\limits_{t_i}^{t_{i+1}}\mathcal{S}^k \mathds{1}(t_{a_k}, t_{b_k} \in [t_i, t_{i+1})), $
\State $\quad \quad \quad \quad \quad \quad \quad \quad\text{s.t.} \quad  0 < t_{1} < t_2 < \cdots < t_{n-1} < 1$
\end{algorithmic}
\label{alg:more_interval}
\end{algorithm*}

\label{apdix:interval}


In this section, we design a procedure to extend \Cref{alg:interval} for case of $n\neq3$. Specifically, we partition the timesteps into $n$ intervals $[0, t_1), \ldots, [t_{n-2}, t_{n-1}), [t_{n-1}, 1]$. Intuitively, the intervals need to minimize the summation of the total functional distance within each partition. This corresponds to the following optimization problem.
%
\begin{equation}
\label{eqn: extended interval opt}
\min_{t_1, \cdots, t_{n+1}} ~ \sum\limits_{k = 0}^{n - 1}\int\limits_{t_k}^{t_{k+1}}\int\limits_{t_k}^{t_{k+1}}\mathcal{S}( \bm \epsilon^*_{t_a},  \bm \epsilon^*_{t_b})\text{d} t_a\text{d} t_b \quad \textrm{s.t.} \quad t_0 = 0, ~t_{n} = 1,~ 0 < t_{1} < t_2 < \cdots < t_{n-1} < 1 \:.
\end{equation}
In order to solve \eqref{eqn: extended interval opt} numerically, we propose \Cref{alg:more_interval}. A solution of the program in step~9 is obtained by discretizeing the time index into $t \in \{0.001, 0.026, \dots, 0.951, 0.976, 1\}$. We use the CIFAR10 dataset with $K = 50000$ to run the experiment. The following table shows the results of different stage numbers $n$ and different split intervals:



%% file: section/appendix_more_exp_arxiv.tex
\section{Additional Experiments}


\label{append:more experiments}

\begin{figure}[t]
    \centering
    \includegraphics[width=0.8\linewidth]{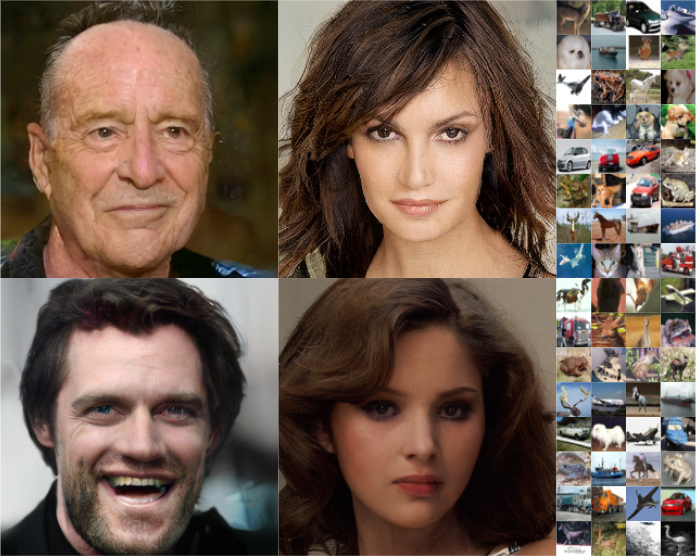}
    \caption{\textbf{Sample generations from Multistage LDM (CelebA $256\times256$) and Multistage DPM-Solver (CIFAR-10 $32\times32$).}}
    \label{fig:visual}
\end{figure}

\subsection{Generation Results}

\label{append:visualization}

The generation samples displayed in \Cref{fig:visual}, from the CelebA and CIFAR-10 datasets, demonstrate that our Multistage strategy is capable of producing high-quality results. Notably, it achieves this with reduced training and computational requirements in comparison to baseline methods.


\subsection{Ablation Study on the Network Parameters}

\label{append:abla_num_parameter}

In this subsection, we perform a series of experiments to determine the best number of parameters within the attempted settings for each stage of our model. For evaluation, we use the best FID scores from all checkpoints. These evaluations are conducted using the DPM-Solver with $20$ Neural Function Evaluations (NFE) on $50,000$ samples. Here, the experimental settings are similar to those used in Section~\ref{sec:exp_setting_evaluation}.


For the Multistage DPM-solver on CIFAR-10 dataset. The models are trained until they achieve their best FID scores. For calculating FID, we follow the methodology described by \cite{song2020denoising}, using the tensorflow\_gan \footnote{\href{https://github.com/tensorflow/gan}{https://github.com/tensorflow/gan}}.


The design of the corresponding parameters for encoders and decoders across different architectures, along with their best FID scores, is detailed in Table \ref{tab:params}. We explore various architectures, aiming to maintain a consistent total number of parameters while varying the ratio of shared parameters (encoder parameters) to the total number of parameters. Our ablation study illustrates that Architecture~1 (our chosen model) indeed exhibits the best FID performance.




\begin{table}[t]
\centering
\begin{tabular}{@{}lllll@{}}
\toprule
Architecture      & Current/Total & Parameters & GFLOPs ($\downarrow$) & FID ($\downarrow$)                   \\ \midrule

\multicolumn{1}{l|}{\multirow{5}{*}{1}} & 1/3                       & 169M   & 29.95  & \multirow{5}{*}{\textbf{2.35}} \\
\multicolumn{1}{l|}{}                   & 2/3                       & 108M   & 17.65  &                       \\
\multicolumn{1}{l|}{}                   & 3/3                       & 47M    & 6.31   &                       \\
\multicolumn{1}{l|}{}                         & Shared                    & 43M    & 5.75   &                       \\
 \multicolumn{1}{l|}{}                        & Total                     & 237M   & 18.65  &                       \\ \midrule
\multicolumn{1}{l|}{\multirow{5}{*}{2}} & 1/3                       & 168M   & 27.57  & \multirow{5}{*}{2.9}  \\
\multicolumn{1}{l|}{}                         & 2/3                       & 138M   & 21.73  &                       \\
\multicolumn{1}{l|}{}                         & 3/3                       & 72M    & 9.65   &                       \\
\multicolumn{1}{l|}{}                         & Shared                    & 68M    & 8.98   &                       \\
\multicolumn{1}{l|}{}                         & Total                     & 242M   & 19.7   &                       \\
                   \midrule
\multicolumn{1}{l|}{\multirow{5}{*}{3}} & 1/3                       & 189M   & 29.43  & \multirow{5}{*}{3.1}  \\
\multicolumn{1}{l|}{}                         & 2/3                       & 160M   & 23.91  &                       \\
\multicolumn{1}{l|}{}                         & 3/3                       & 103M   & 13.7   &                       \\
\multicolumn{1}{l|}{}                         & Shared                    & 98M    & 12.93  &                       \\
\multicolumn{1}{l|}{}                         & Total                     & 256M   & 22.45  &                       \\
                   \midrule
\multicolumn{1}{l|}{\multirow{5}{*}{4}} & 1/3                       & 174M   & 26.54  & \multirow{5}{*}{3.34} \\
 \multicolumn{1}{l|}{}                        & 2/3                       & 126M   & 17.62  &                       \\
 \multicolumn{1}{l|}{}                        & 3/3                       & 103M   & 13.7   &                       \\
  \multicolumn{1}{l|}{}                       & Shared                    & 98M    & 12.93  &                       \\
 \multicolumn{1}{l|}{}                        & Total                     & 207M   & 22.45  &                       \\
\bottomrule
\end{tabular}
\caption{\textbf{Ablation study of Multistage DPM-solver on CIFAR-10 dataset.} Current/Total: the number of the stage over the total number of stages. Shared: the encoder. Total: combined the encoder and decoders. Total GFLOPS: averaged GFLOPS of all stages weighted by NFE assigned to each stage during sampling of DPM-Solver.}
    \label{tab:params}
\end{table}
For Latent Diffusion multistage models on CelebA-HQ, models are trained for 500k iterations, and the best checkpoints are used for evaluation. We compute the FID following \cite{rombach2022high} using the \texttt{torch-fidelity} package \cite{torchfid}. The parameter design and best FID are shown in \Cref{tab:ldmParams}. Our ablation study shows the superiority of utilizing Architecture~4 (our chosen model) in terms of the reported FID values. 
\begin{table}[t]
\centering
\begin{tabular}{@{}lllll@{}}
\toprule
Architecture & Current/Total & Parameters & GFLOPs ($\downarrow$) & FID ($\downarrow$)      \\ \midrule
\multicolumn{1}{l|}{\multirow{5}{*}{1}} & 1/3 & 238M & 80.03 & \multirow{5}{*}{9.37}\\                      \multicolumn{1}{l|}{} &2/3 & 189M   & 64.95   &      \\
\multicolumn{1}{l|}{} &3/3  & 161M   & 51.87  &  \\
\multicolumn{1}{l|}{} & Shared   & 79M     & 16.69  &    \\
\multicolumn{1}{l|}{} & Total                   & 428M                         & 65.75                      & 
 \\ \midrule
\multicolumn{1}{l|}{\multirow{5}{*}{2}} & 1/3 & 206M & 72.61    &   \multirow{5}{*}{9.73}                          \\
\multicolumn{1}{l|}{} & 2/3 & 158M                         & 57.85                      &                            \\
\multicolumn{1}{l|}{}  & 3/3 & 108M                         & 34.38                      &                            \\
\multicolumn{1}{l|}{}  & Shared                  & 55M                          & 11.6                       &                            \\
\multicolumn{1}{l|}{}                         & Total                   & 361M                         & 54.37                    & \\ \midrule
\multicolumn{1}{l|}{\multirow{5}{*}{3}} & 1/3 & 274M & 88.39                      & \multirow{5}{*}{8.43}                           \\
\multicolumn{1}{l|}{} & 2/3 & 224M                         & 72.97                      &                            \\
\multicolumn{1}{l|}{} &3/3 & 170M                         & 48.17                      &                            \\
\multicolumn{1}{l|}{} & Shared                  & 108M                         & 22.71                      &                            \\
\multicolumn{1}{l|}{} & Total                & 452M                         & 69.22                     &      \\ \midrule
\multicolumn{1}{l|}{\multirow{5}{*}{4}} & 1/3 & 316M & 105.82                     & \multirow{5}{*}{\textbf{8.38}}                            \\
\multicolumn{1}{l|}{} & 2/3 & 224M                         & 72.97                      &                            \\
\multicolumn{1}{l|}{} &3/3 & 170M                         & 48.17                      &                            \\
\multicolumn{1}{l|}{} & Shared                  & 108M                         & 22.71                      &                            \\
\multicolumn{1}{l|}{} & Total                    & 494M                         & 76.19                      &     \\ \bottomrule
\end{tabular}
\caption{\textbf{Ablation study of Multistage LDM on CelebA dataset.}}
    \label{tab:ldmParams}
\end{table}

\subsection{Ablation Study on the Number of Intervals}
\label{append:abla_num_interval}

In this subsection, we conduct ablation studies on the number of intervals of our multistage architecture. We utilize Multistage DPM-Solver training on the CIFAR-10 dataset. To extend \Cref{alg:interval}, we design the procedure outlined in \Cref{alg:more_interval} (\Cref{append:more interval}). The splitting intervals and results are given in \Cref{tab:numstages}. 
We compare the stages from one to five. The parameters designed for those architectures are based on the best three-stage model architecture in \Cref{tab:params}. More specifically, we use linear interpolation. Suppose the partitioning timesteps for the three-stage architecture are $t_1$ and $t_2$ and parameter numbers for each interval are $p_1, p_2$, and $p_3$. The parameter $p$ for a random interval $[t_a, t_b]$ is calculated as:
\begin{equation}
    \label{eqn: pp}
    p = \frac{1}{L ([t_a, t_b]) } [\mathcal L ([t_a, t_b] \cap [0, t_1)) p_1 ~ + ~    \mathcal L ([t_a, t_b] \cap [t_1, t_2)) p_2 ~+~ \mathcal L ([t_a, t_b] \cap [t_2, 1]) p_3 ]\:,
\end{equation}
where $\mathcal L$ is an operator that measures the length of an interval. We design the parameters for all architectures in \Cref{tab:numstages} based on \eqref{eqn: pp}, and split intervals for each stage accordingly. As observed, the three-stage architecture, adopted in the main body of the paper, outperforms all other settings.

\begin{table}[t]
\centering
\begin{tabular}{@{}llllll@{}}
\toprule
Number of Stages  & Current/Total & Interval & Parameters & GFLOPs ($\downarrow$) & FID ($\downarrow$)                 \\ \midrule
\multicolumn{1}{l|}{1}                     & 1/1  & [0,1]         & 108M   & 17.65  & 2.75                  \\  \midrule
\multicolumn{1}{l|}{\multirow{2}{*}{2}}    & 1/2  & [0,0.476)         & 136M   & 26.59  & \multirow{4}{*}{2.56} \\
\multicolumn{1}{l|}{}                                     & 2/2    & (0.476,1]       & 95M    & 16.85  &                       \\
\multicolumn{1}{l|}{}                                     & Total  &        & 188M   & 21.49  &                       \\
\midrule
\multicolumn{1}{l|}{\multirow{3}{*}{3}} & 1/3  & [0,0.442)         & 169M   & 29.95  & \multirow{5}{*}{\textbf{2.35}} \\
\multicolumn{1}{l|}{}                                     & 2/3   & [0.442, 0.630)        & 108M   & 17.65  &                       \\
\multicolumn{1}{l|}{}                                     & 3/3  & [0.630,1]         & 47M    & 6.31   &                       \\
\multicolumn{1}{l|}{}                                     & Total &        & 230M   & 18.65  &                       \\ \midrule
\multicolumn{1}{l|}{\multirow{4}{*}{4}}                   & 1/4 & [0, 0.376)          & 159M   & 29.78  & \multirow{6}{*}{2.67} \\
\multicolumn{1}{l|}{}                                     & 2/4 & [0.376, 0.526)          & 95M    & 16.85  &                       \\
\multicolumn{1}{l|}{}                                     & 3/4 & [0.526, 0.726)          & 71M    & 11.97  &                       \\
\multicolumn{1}{l|}{}                                     & 4/4 & [0.726, 1]          & 32M    & 4.41   &                       \\
\multicolumn{1}{l|}{}                                     & Total &        & 284M   & 17.33  &                       \\
\midrule
\multicolumn{1}{l|}{\multirow{5}{*}{5}}                   & 1/5  & [0, 0.376)         & 154M   & 30.38  & \multirow{7}{*}{2.88} \\
\multicolumn{1}{l|}{}                                     & 2/5  & [0.376, 0.476)         & 88M    & 16.82  &                       \\
\multicolumn{1}{l|}{}                                     & 3/5  & [0.476, 0.626)         & 88M    & 16.82  &                       \\
\multicolumn{1}{l|}{}                                     & 4/5  & [0.626, 0.776)         & 27M    & 4.42   &                       \\
\multicolumn{1}{l|}{}                                     & 5/5  & [0.776, 1]         & 21M    & 3.28   &                       \\
\multicolumn{1}{l|}{}                                     & Total  &       & 336M   & 17.03  & \\  \bottomrule                    
\end{tabular}
\caption{\textbf{Ablation study on the number of stages.}}
    \label{tab:numstages}
\end{table}

\subsection{Experimental Setting on Interval Splitting Methods}

\label{append:exp_set_interval_split}

This section presents the experimental settings of previous timestep clustering algorithms considered in \Cref{tab:cluster}. We refer readers to the following references \cite{go2023addressing,lee2023multi} for implementation details. Specifically, we compare these methods for the 3 intervals case on CIFAR-10 dataset by training our proposed Multistage architecture with calculated intervals until convergence. The timestep and SNR-based clustering algorithms are relatively easy to implement. For the gradient-based clustering, we collect the gradients of the trained network by the DPM-Solver for clustering. These gradients are generated every 50k training iterations with a total of 450k iterations. We evaluate the performances of these models based on their best FID scores across training iterations.

%% file: main.bib
@String(ICCV= {Int. Conf. Comput. Vis.})

@String(ICCV  = {ICCV})

@article{karras2022elucidating,
  title={Elucidating the design space of diffusion-based generative models},
  author={Karras, Tero and Aittala, Miika and Aila, Timo and Laine, Samuli},
  journal={arXiv preprint arXiv:2206.00364},
  year={2022}
}

@article{song2020score,
  title={Score-based generative modeling through stochastic differential equations},
  author={Song, Yang and Sohl-Dickstein, Jascha and Kingma, Diederik P and Kumar, Abhishek and Ermon, Stefano and Poole, Ben},
  journal={arXiv preprint arXiv:2011.13456},
  year={2020}
}

@article{lu2022dpm,
  title={Dpm-solver: A fast ode solver for diffusion probabilistic model sampling in around 10 steps},
  author={Lu, Cheng and Zhou, Yuhao and Bao, Fan and Chen, Jianfei and Li, Chongxuan and Zhu, Jun},
  journal={arXiv preprint arXiv:2206.00927},
  year={2022}
}

@inproceedings{ronneberger2015u,
  title={U-net: Convolutional networks for biomedical image segmentation},
  author={Ronneberger, Olaf and Fischer, Philipp and Brox, Thomas},
  booktitle={Medical Image Computing and Computer-Assisted Intervention--MICCAI 2015: 18th International Conference, Munich, Germany, October 5-9, 2015, Proceedings, Part III 18},
  pages={234--241},
  year={2015},
  organization={Springer}
}

@inproceedings{choi2022perception,
  title={Perception prioritized training of diffusion models},
  author={Choi, Jooyoung and Lee, Jungbeom and Shin, Chaehun and Kim, Sungwon and Kim, Hyunwoo and Yoon, Sungroh},
  booktitle={Proceedings of the IEEE/CVF Conference on Computer Vision and Pattern Recognition},
  pages={11472--11481},
  year={2022}
}

@article{alkhouri2023diffusion,
  title={Diffusion-based Adversarial Purification for Robust Deep MRI Reconstruction},
  author={Alkhouri, Ismail and Liang, Shijun and Wang, Rongrong and Qu, Qing and Ravishankar, Saiprasad},
  journal={arXiv preprint arXiv:2309.05794},
  year={2023}
}

@article{go2023addressing,
  title={Addressing Negative Transfer in Diffusion Models},
  author={Go, Hyojun and Kim, JinYoung and Lee, Yunsung and Lee, Seunghyun and Oh, Shinhyeok and Moon, Hyeongdon and Choi, Seungtaek},
  journal={arXiv preprint arXiv:2306.00354},
  year={2023}
}

@article{krizhevsky2009learning,
  title={Learning multiple layers of features from tiny images},
  author={Krizhevsky, Alex and Hinton, Geoffrey and others},
  year={2009},
  publisher={Toronto, ON, Canada}
}

@article{ho2020denoising,
  title={Denoising diffusion probabilistic models},
  author={Ho, Jonathan and Jain, Ajay and Abbeel, Pieter},
  journal={Advances in neural information processing systems},
  volume={33},
  pages={6840--6851},
  year={2020}
}

@article{lee2023multi,
  title={Multi-Architecture Multi-Expert Diffusion Models},
  author={Lee, Yunsung and Kim, Jin-Young and Go, Hyojun and Jeong, Myeongho and Oh, Shinhyeok and Choi, Seungtaek},
  journal={arXiv preprint arXiv:2306.04990},
  year={2023}
}

@inproceedings{go2023towards,
  title={Towards practical plug-and-play diffusion models},
  author={Go, Hyojun and Lee, Yunsung and Kim, Jin-Young and Lee, Seunghyun and Jeong, Myeongho and Lee, Hyun Seung and Choi, Seungtaek},
  booktitle={Proceedings of the IEEE/CVF Conference on Computer Vision and Pattern Recognition},
  pages={1962--1971},
  year={2023}
}

@article{balaji2022ediffi,
  title={ediffi: Text-to-image diffusion models with an ensemble of expert denoisers},
  author={Balaji, Yogesh and Nah, Seungjun and Huang, Xun and Vahdat, Arash and Song, Jiaming and Kreis, Karsten and Aittala, Miika and Aila, Timo and Laine, Samuli and Catanzaro, Bryan and others},
  journal={arXiv preprint arXiv:2211.01324},
  year={2022}
}

@article{zhang2022gddim,
  title={gDDIM: Generalized denoising diffusion implicit models},
  author={Zhang, Qinsheng and Tao, Molei and Chen, Yongxin},
  journal={arXiv preprint arXiv:2206.05564},
  year={2022}
}

@inproceedings{he2017mask,
  title={Mask r-cnn},
  author={He, Kaiming and Gkioxari, Georgia and Doll{\'a}r, Piotr and Girshick, Ross},
  booktitle={Proceedings of the IEEE international conference on computer vision},
  pages={2961--2969},
  year={2017}
}

@article{heusel2017gans,
  title={Gans trained by a two time-scale update rule converge to a local nash equilibrium},
  author={Heusel, Martin and Ramsauer, Hubert and Unterthiner, Thomas and Nessler, Bernhard and Hochreiter, Sepp},
  journal={Advances in neural information processing systems},
  volume={30},
  year={2017}
}

@inproceedings{liu2015faceattributes,
  title = {Deep Learning Face Attributes in the Wild},
  author = {Liu, Ziwei and Luo, Ping and Wang, Xiaogang and Tang, Xiaoou},
  booktitle = {Proceedings of International Conference on Computer Vision (ICCV)},
  year = {2015} 
}

@article{song2020denoising,
  title={Denoising diffusion implicit models},
  author={Song, Jiaming and Meng, Chenlin and Ermon, Stefano},
  journal={arXiv preprint arXiv:2010.02502},
  year={2020}
}

@article{zhang2022fast,
  title={Fast Sampling of Diffusion Models with Exponential Integrator},
  author={Zhang, Qinsheng and Chen, Yongxin},
  journal={arXiv preprint arXiv:2204.13902},
  year={2022}
}

@article{xu2022poisson,
  title={Poisson flow generative models},
  author={Xu, Yilun and Liu, Ziming and Tegmark, Max and Jaakkola, Tommi},
  journal={Advances in Neural Information Processing Systems},
  volume={35},
  pages={16782--16795},
  year={2022}
}

@article{dhariwal2021diffusion,
  title={Diffusion models beat gans on image synthesis},
  author={Dhariwal, Prafulla and Nichol, Alexander},
  journal={Advances in neural information processing systems},
  volume={34},
  pages={8780--8794},
  year={2021}
}

@article{ho2022classifier,
  title={Classifier-free diffusion guidance},
  author={Ho, Jonathan and Salimans, Tim},
  journal={arXiv preprint arXiv:2207.12598},
  year={2022}
}

@article{su2022dual,
  title={Dual diffusion implicit bridges for image-to-image translation},
  author={Su, Xuan and Song, Jiaming and Meng, Chenlin and Ermon, Stefano},
  journal={arXiv preprint arXiv:2203.08382},
  year={2022}
}

@inproceedings{saharia2022palette,
  title={Palette: Image-to-image diffusion models},
  author={Saharia, Chitwan and Chan, William and Chang, Huiwen and Lee, Chris and Ho, Jonathan and Salimans, Tim and Fleet, David and Norouzi, Mohammad},
  booktitle={ACM SIGGRAPH 2022 Conference Proceedings},
  pages={1--10},
  year={2022}
}

@article{zhao2022egsde,
  title={Egsde: Unpaired image-to-image translation via energy-guided stochastic differential equations},
  author={Zhao, Min and Bao, Fan and Li, Chongxuan and Zhu, Jun},
  journal={Advances in Neural Information Processing Systems},
  volume={35},
  pages={3609--3623},
  year={2022}
}

@inproceedings{rombach2022high,
  title={High-resolution image synthesis with latent diffusion models},
  author={Rombach, Robin and Blattmann, Andreas and Lorenz, Dominik and Esser, Patrick and Ommer, Bj{\"o}rn},
  booktitle={Proceedings of the IEEE/CVF conference on computer vision and pattern recognition},
  pages={10684--10695},
  year={2022}
}

@inproceedings{ramesh2021zero,
  title={Zero-shot text-to-image generation},
  author={Ramesh, Aditya and Pavlov, Mikhail and Goh, Gabriel and Gray, Scott and Voss, Chelsea and Radford, Alec and Chen, Mark and Sutskever, Ilya},
  booktitle={International Conference on Machine Learning},
  pages={8821--8831},
  year={2021},
  organization={PMLR}
}

@article{nichol2021glide,
  title={Glide: Towards photorealistic image generation and editing with text-guided diffusion models},
  author={Nichol, Alex and Dhariwal, Prafulla and Ramesh, Aditya and Shyam, Pranav and Mishkin, Pamela and McGrew, Bob and Sutskever, Ilya and Chen, Mark},
  journal={arXiv preprint arXiv:2112.10741},
  year={2021}
}

@article{song2023solving,
  title={Solving Inverse Problems with Latent Diffusion Models via Hard Data Consistency},
  author={Song, Bowen and Kwon, Soo Min and Zhang, Zecheng and Hu, Xinyu and Qu, Qing and Shen, Liyue},
  journal={arXiv preprint arXiv:2307.08123},
  year={2023}
}

@article{chung2022diffusion,
  title={Diffusion posterior sampling for general noisy inverse problems},
  author={Chung, Hyungjin and Kim, Jeongsol and Mccann, Michael T and Klasky, Marc L and Ye, Jong Chul},
  journal={arXiv preprint arXiv:2209.14687},
  year={2022}
}

@article{song2021solving,
  title={Solving inverse problems in medical imaging with score-based generative models},
  author={Song, Yang and Shen, Liyue and Xing, Lei and Ermon, Stefano},
  journal={arXiv preprint arXiv:2111.08005},
  year={2021}
}

@article{ho2022imagen,
  title={Imagen video: High definition video generation with diffusion models},
  author={Ho, Jonathan and Chan, William and Saharia, Chitwan and Whang, Jay and Gao, Ruiqi and Gritsenko, Alexey and Kingma, Diederik P and Poole, Ben and Norouzi, Mohammad and Fleet, David J and others},
  journal={arXiv preprint arXiv:2210.02303},
  year={2022}
}

@article{harvey2022flexible,
  title={Flexible diffusion modeling of long videos},
  author={Harvey, William and Naderiparizi, Saeid and Masrani, Vaden and Weilbach, Christian and Wood, Frank},
  journal={Advances in Neural Information Processing Systems},
  volume={35},
  pages={27953--27965},
  year={2022}
}

@article{song2023consistency,
  title={Consistency models},
  author={Song, Yang and Dhariwal, Prafulla and Chen, Mark and Sutskever, Ilya},
  year={2023}
}

@article{vincent2011connection,
  title={A connection between score matching and denoising autoencoders},
  author={Vincent, Pascal},
  journal={Neural computation},
  volume={23},
  number={7},
  pages={1661--1674},
  year={2011},
  publisher={MIT Press}
}

@misc{torchfid,
  author={Anton Obukhov and Maximilian Seitzer and Po-Wei Wu and Semen Zhydenko and Jonathan Kyl and Elvis Yu-Jing Lin},
  year=2020,
  title={High-fidelity performance metrics for generative models in PyTorch},
  url={https://github.com/toshas/torch-fidelity},
  publisher={Zenodo},
  version={v0.2.0},
  doi={10.5281/zenodo.3786540},
  note={Version: 0.2.0, DOI: 10.5281/zenodo.3786540}
}

@article{zhang2023emergence,
  title={The emergence of reproducibility and consistency in diffusion models},
  author={Zhang, Huijie and Zhou, Jinfan and Lu, Yifu and Guo, Minzhe and Shen, Liyue and Qu, Qing},
  journal={arXiv preprint arXiv:2310.05264},
  year={2023}
}

@inproceedings{yang2023skypilot,
  title={$\{$SkyPilot$\}$: An Intercloud Broker for Sky Computing},
  author={Yang, Zongheng and Wu, Zhanghao and Luo, Michael and Chiang, Wei-Lin and Bhardwaj, Romil and Kwon, Woosuk and Zhuang, Siyuan and Luan, Frank Sifei and Mittal, Gautam and Shenker, Scott and others},
  booktitle={20th USENIX Symposium on Networked Systems Design and Implementation (NSDI 23)},
  pages={437--455},
  year={2023}
}
